\documentclass{article}

\usepackage{arxiv}

\usepackage[utf8]{inputenc} 
\usepackage[T1]{fontenc}    
\usepackage{hyperref}       
\usepackage{url}            
\usepackage{booktabs}       
\usepackage{amsfonts}       
\usepackage{amsmath}
\usepackage{nicefrac}       
\usepackage{microtype}      
\usepackage{graphicx}
\usepackage{natbib}
\usepackage{doi}
\usepackage[onehalfspacing]{setspace}
\usepackage{xcolor}
\usepackage{subcaption}
\usepackage{graphicx}

\begin{document}

\title{
Human-in-the-loop: Towards Label Embeddings for Measuring Classification Difficulty 
}

\author{Katharina Hechinger$^{*}$ \\
	Department of Statistics\\
	Ludwig-Maximilians-University\\
	Munich, Germany \\
 \And
  Christoph Koller  \\
 Chair of Data Science in Earth Observation \\
	Technical University of Munich\\
	   Munich, Germany \\
    German Aerospace Center \\
    Wessling, Germany
 \And
 Xiao Xiang Zhu \\
	Chair of Data Science in Earth Observation \\
	Technical University of Munich\\
	Munich, Germany\\
    Munich Center for Machine Learning \\
    Munich, Germany
 \And
   G\"{o}ran Kauermann \\
        Department of Statistics\\
        Ludwig-Maximilians-University\\
	Munich, Germany \\
}

\maketitle
\def\thefootnote{*}\footnotetext{\texttt{katharina.hechinger@stat.uni-muenchen.de}}

\begin{abstract}
\small

Uncertainty in machine learning models is a timely and vast field of research. In supervised learning, uncertainty can already occur in the first stage of the training process, the annotation phase.
This scenario is particularly evident when some instances cannot be definitively classified. In other words, there is inevitable ambiguity in the annotation step and hence, not necessarily a single "ground truth" associated with each instance.  


The main idea of this work is to drop the assumption of a ground truth label and instead embed the annotations into a multidimensional space. This embedding is derived from the empirical distribution of annotations in a Bayesian setup, modeled via a Dirichlet-Multinomial framework. We estimate the model parameters and posteriors using a stochastic Expectation Maximisation algorithm with Markov Chain Monte Carlo steps. 
The methods developed in this paper readily extend to various situations where multiple annotators independently label instances. To showcase the generality of the proposed approach, we apply our approach to three benchmark datasets for image classification and Natural Language Inference, where multiple annotations per instance are available. Besides the embeddings, we can investigate the resulting correlation matrices, which reflect the semantic similarities of the original classes very well for all three exemplary datasets.

\end{abstract}


\section{Introduction}
\label{sec:intro}

Machine Learning models are increasingly used for a growing number of applications, one of which is supervised classification, for example in the form of images or texts. While such models have achieved impressive standards over the last years in terms of accuracy, the assessment of uncertainty remains an active field of open problems and research challenges. Recent survey articles discussing the field include \citet{gawlikowski:2021} or \citet{hullermeier:2021}. Uncertainty thereby has numerous and intertwined sources as discussed in \citet{gruber:2023} or \citet{baan:2023} and is heavily impacted at multiple stages of the common machine learning pipeline. \citet{gruber:2023} also explicitly emphasize the role of the data itself for appropriately assessing uncertainty entirely.

In the field of deep learning, multiple major streams of research related to the quantification of uncertainty exist. Besides ensemble methods and Bayesian approaches, evidential neural networks have been gaining attention as a deterministic method of uncertainty quantification (\citealp{sensoy:2018}). Specifically, these methods conceptualize learning as the acquisition of evidence, where each new training example adds support to a learned evidential distribution.
A recent survey by \cite{ulmer:2023} provides an extensive overview of evidential deep learning and discusses its strengths and weaknesses in depth.
However, some lines of work also advise caution when employing evidential networks for uncertainty quantification. \cite{jurgens:2024} state that generally, epistemic uncertainty is not reliably represented by those methods and \cite{meinert:2023} showcase the issue of overparameterization for evidential regression.

However, while some parts of the overall uncertainty are already heavily studied in research, less attention has been paid to one of the major prerequisites for training classification models, namely the (un-)availability of reliable ground truth labels for the training data and their uncertainty. In fact, uncertainty already starts in the labeling process for supervised machine learning, where human annotators label images or texts. Any supervised model will rely on these ``ground truth'' labels, inherently incorporating their associated uncertainty. We refer to this type of uncertainty as ``label uncertainty''. Commonly, such gold labels are acquired with human labeling effort, leading to multiple annotations per instance. Depending on the complexity of the problem at hand, it might suffice to aggregate the annotations into a single ground truth label, e.g. by majority voting. However, in many realistic application areas, such as the classification of complex images or the assessment of language and speech, this assumption does not hold true. \\
Of course, humans are naturally prone to errors, leading to unreliable annotations or mistakes, and therewith, label noise or label errors.
The problem was already tackled and discussed early in the statistical literature, see for example \citet{dawid:1979}. In recent years, more and more methods have been developed for handling data despite human errors, for example in the context of neural networks (\citealp{dgani:2018}). \cite{peterson:2019} argue that incorporating human ambiguity can improve classification models in terms of robustness. However, training supervised machine learning models based on noisy or deficient labels can lead to poor performance and high uncertainties, see e.g. \citet{frenay:2014} or more recently \citet{frenay:2021} for an overview. Also, the labels might introduce some bias, as shown by \citet{jiang:2020}, that needs to be identified and corrected if possible. Different algorithms have been introduced to tackle the problem of noisy labels, see \citet{algan:2021} for an extensive survey on various methods. 

However, ambiguity in annotations cannot always be attributed to the fallibility of human annotators. Instead, label variation is also likely to arise if the assumption of a singular ground truth label for each instance is questionable. 
In the context of language, \citet{plank:2022} discusses the sources of label variation. Particularly, the authors argue that the absence of a singular ground truth is often reasonable and should not be considered erroneous by default. In this line, the survey by \citet{uma:2021} discusses the disagreement of annotators. The authors conclude that suitable evaluation methods are required if a single gold label cannot be assigned. Various works in multiple domains show that disregarding label variation and leaving it untreated can indeed lead to quality issues and uncertainties. The common approach to simply summarise the annotations into a single label does not only discard valuable information, it is also an inappropriate representation of the truth and introduces remarkable amounts of uncertainty, in particular in the ``gold'' label (\citealp{davani:2022}, \citealp{uma:2021} or \citealp{aroyo:2013}).

\begin{figure}[t]
    \centering
    \includegraphics[width=0.8\textwidth]{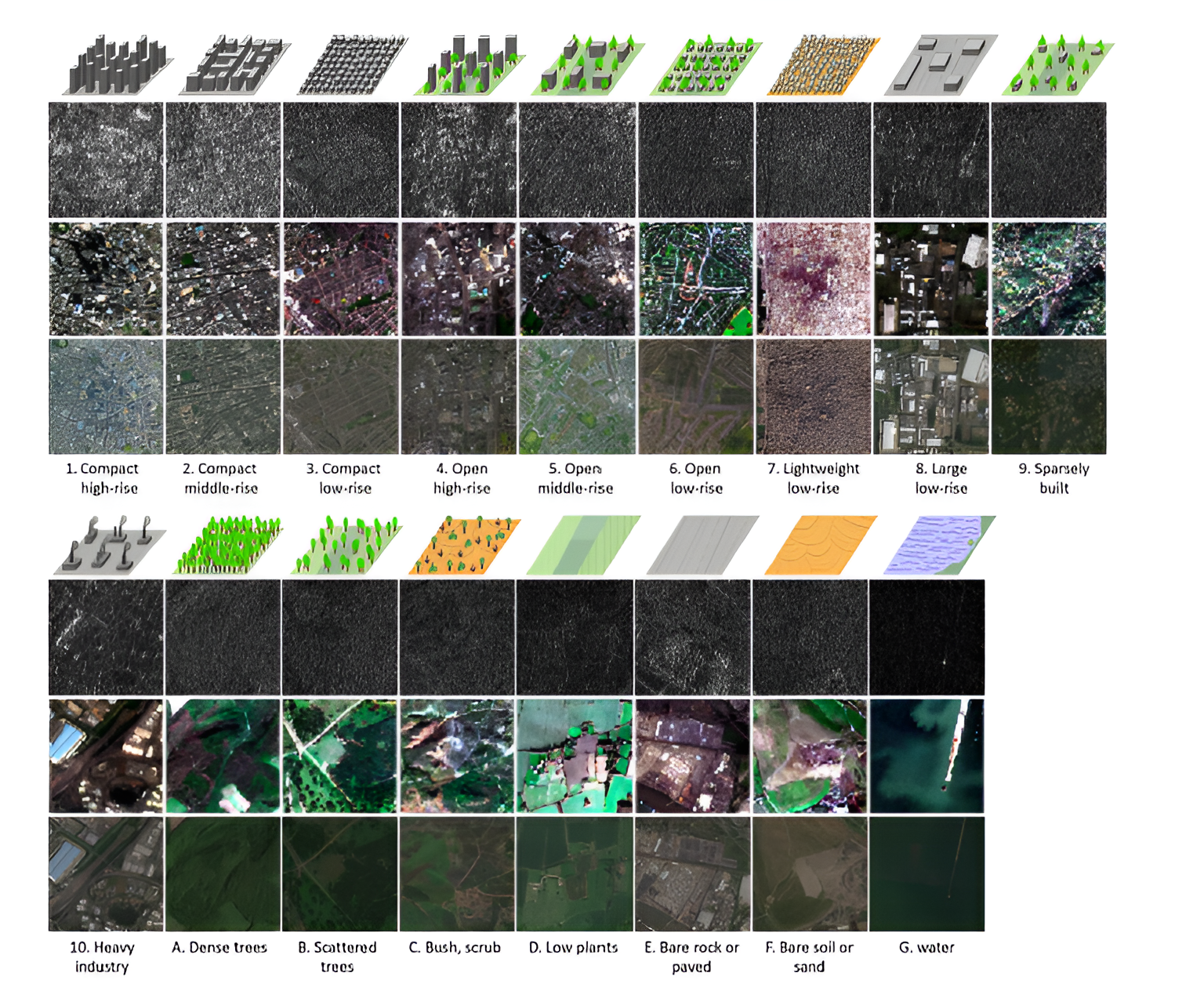}
    \caption{The figure shows the categories of the LCZ classification scheme (\citealp{Stewart:2012}), along with exemplary images for each class from three different sources (top row: Sentinel 1, middle row: Sentinel 2, bottom row: Google Earth), images taken from \cite{Zhu:2020}.}
    \label{fig:intro_lcz_scheme}
\end{figure}

\begin{figure}[t]
\centering

\begin{subfigure}{0.23\textwidth}
\centering
    \includegraphics{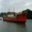}
    \caption{\footnotesize{Original label: Boat}}
    \label{fig:third}
\end{subfigure}
\hfill
\begin{subfigure}{0.23\textwidth}
\centering
    \includegraphics{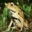}
    \caption{\footnotesize{Original label: Frog}}
    \label{fig:third}
\end{subfigure}
\hfill
\begin{subfigure}{0.23\textwidth}
    \centering
    \includegraphics{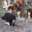}
    \caption{\footnotesize{Original label: Cat}}
    \label{fig:first}
\end{subfigure}
\hfill
\begin{subfigure}{0.23\textwidth}
\centering
    \includegraphics{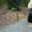}
    \caption{\footnotesize{Original label: Deer}}
    \label{fig:second}
\end{subfigure}
        
\caption{The figure shows exemplary images from Cifar-10H, where a high disagreement rate between the annotators could be observed hinting at the ambiguity of the images.}
\label{fig:intro_examples}
\end{figure}

This problem is prevalent across various classification domains, specifically for application areas characterized by inherent ambiguity. To showcase this, 
let us first consider the domain of natural language processing (NLP) or more specifically natural language inference (NLI), where ambiguity is ubiquitous due to the subjective interpretation of language and speech.
This issue has been already extensively discussed, see e.g.\ \cite{plank:2022}. NLI corresponds to the task of discerning the logical relationship between two sentences, typically whether one entails the other, contradicts it, or is unrelated to it. Naturally, the perception of language differs for the human annotators causing high rates of disagreement (\citealp{nie:2020}, \citealp{pavlick:2019}). Table \ref{tab:intro_examples} shows exemplary sentences, which are highly ambiguous. This ambiguity is clearly reflected by the annotations. \cite{gruber:2024} provide a statistical approach for modeling the data-generating process in order to gain a better understanding of the label uncertainty. However, their modeling approach assumes a latent ground truth label associated with each sentence pair. While this is only a modeling assumption, numerous works claim that the assumption of a single ground truth is not appropriate for NLI tasks and instead, a more realistic representation of the labels should be used, see e.g.\ \cite{aroyo:2015}, \cite{uma:2021} or \cite{plank:2022}. \\
Similar problems arise in the domain of image classification if either the categories or the images themselves are ambiguous. Depending on the nature of the problem, assigning a singular ground truth label is often simply impossible. 
Our second example comes from  the field of remote sensing and satellite image classification. The ultimate goal is to categorize images into local climatic zones (LCZs), a classification scheme for satellite images developed by \cite{Stewart:2012}. A huge effort has been made to develop complex deep neural networks for this task (e.g., \citealp{Zhu:2020}, \citealp{Qiu:2019} or \citealp{Qiu:2020}). 
These models rely on large amounts of labeled data and, hence, it is necessary to manually annotate a vast amount of satellite images. Annotation of such images requires specialized expertise in earth observation and hence, the labeling is conducted by trained experts instead of laypersons. However, even with their domain knowledge, annotators frequently disagree due to the complexity of the task and the inherent ambiguity of both, the images and the categories. Figure \ref{fig:intro_lcz_scheme} illustrates the climatic classes with accompanying low-resolution satellite images and high-resolution images from Google Earth, highlighting the challenges in classification and the potential for disagreement.
Nevertheless, the annotations are commonly aggregated into a majority vote discarding valuable information about the initial disagreement (\citealp{Zhu:2020}), and the uncertainty associated with the ground truth labels remains untreated. \citet{hechinger:2024} approach this problem from a statistical perspective and employ a Bayesian mixture model to estimate the latent posterior label distributions for the images based on the annotations. Based on the proposed modeling framework they assess factors contributing to the uncertainty and variation in the annotations. Their work especially showcases the heterogeneity of the individual annotators and the diversity of the images themselves. These factors lead to remarkable variation within the annotations and, hence, express the complexity of the labeling task. Their work is again based on the modeling assumption of a singular ground truth label for each image. However, the validity of this assumption in practice is questionable due to the ambiguous nature of the images themselves. \\
Lastly, situations without a distinct ground truth might also arise for presumably straightforward classification tasks, as shown in Figure \ref{fig:intro_examples}. The exemplary images are part of the benchmark dataset Cifar-10H, introduced by \cite{peterson:2019}. The dataset is designed such that each instance is assigned to a single unambiguous class, at least in theory. 
Still, some images defy easy classification due to ambiguities caused by the size or quality of the picture, leading to high disagreement rates within the annotations. Consequently, relying solely on majority voting to assign a singular label in such cases does not accurately reflect the underlying truth. We take this data set as a third example to showcase our modeling approach. 

\begin{table}
    \centering
    \footnotesize
    \begin{tabular}{lll}
    & & \textbf{Human Votes} \\
        \textbf{Context/Premise} & \textbf{Statement/Hypothesis} &  \textbf{[C, N, E]}  \\ \hline 
          A man running a marathon talks to his friend. & There is a man running. & [0, 0, 100] \\
         A black and white dog running through shallow water. &  Two dogs running through water. & [42, 14, 44] \\
         A woman holding a child in a purple shirt. & The woman is asleep at home. & [46, 53, 1] \\
         An elderly woman crafts a design on a loom. & The woman is sewing. & [34, 31, 35] \\
         &&
    \end{tabular}
    \caption{The table shows 4 examples of sentence pairs from ChaosSNLI, along with the annotations, see \cite{gruber:2024}. Each pair of context and statement is classified by 100 human annotators with the categories ``contradiction'' (C), ``neutral'' (N) and ``entailment'' (E). }
    \label{tab:intro_examples}
\end{table}

In this work, we propose to move away from the premise of a sole ground truth. The three examples sketched above underpin that the assumption of a ground truth is not always appropriate. Therefore, we explicitly allow for ambiguity for each instance. This view extends to many real-world applications, where it is likely to encounter instances that can not be classified uniquely but are associated with a mixture or combination of different classes. In this paper, we aim to statistically model such situations in a distributional framework. Namely, we employ a Dirichlet Multinomial model, as discussed in \cite{minka:2000} or \citet{mosimann:1962}. Variants of this model class have been e.g.\ used for clustering of text documents (\citealp{yin:2014}) or genomics data (\citealp{holmes:2012} or \citealp{harrison:2020}). \citet{avetisyan:2012} deploy a Dirichlet-Multinomial Mixture model to estimate survey response rates. \citet{eswaran:2017} also connected this model class to uncertainty quantification and modeled beliefs as Dirichlet distributions to capture uncertainty. 
In this work, we propose a Dirichlet Multinomial model to estimate \textbf{embedded ground truth values} that express the classification difficulty and uncertainty for the respective images based on human annotations. Specifically, we construct an embedding space so that each image is located in a $K$ dimensional space, with $K$ as the number of categories. To do so, we pursue an empirical Bayes approach in combination with Markov Chain Monte Carlo (MCMC) sampling and a stochastic version of the Expectation Maximization (EM) algorithm for estimation, as proposed by \cite{Celeux:1996}. The presented strategy gives insights into the correlation (or confusion) patterns between different classes and simultaneously allows to express and quantify uncertainty. 

Moving forward, the results can subsequently be integrated into the machine learning pipeline by directly training a model on the acquired embedded labels rather than relying on the majority-voted classes. 
Approaches in this direction are based on the idea of label distribution learning, which incorporates the ambiguity of labels (e.g. \citealp{geng:2016}, \citealp{gao:2017} or \citealp{xu:2019}). \citet{koller:2024} suggest integrating the label uncertainty into the training process by using distributional labels based on multiple votes for better generalization for unseen data and more stable performance in terms of uncertainty calibration. Extending their ideas by using a more sound representation of the labels, i.e. the label embeddings estimated by our approach, could allow us to improve the predictions even more in terms of uncertainty. We will sketch the idea but this paper emphasizes the statistical modelling step and not on the incorporation into a machine learning pipeline. 

The paper is structured as follows. Section 2 describes the distributional framework and the algorithm used for estimation. First, we consider a binary case with two classes only for clarification purposes and then move on to the more general multiclass case. The results on three different datasets are reported in Section 3. We consider some possible further steps and applications in Section 4. Section 5 concludes the paper with a detailed discussion.

\section{Model}
\label{sec:model}

\subsection{Notations}

Each image $i$, with $i=1,...,n$ is assessed by a set of annotators (labelers, voters) indexed with $j$, where $j=1,...,J_i$. We consider the images as independent and the same holds for the annotators. The labelers classify each image individually into the class $k$, where  $k=1,...,K$. The corresponding vote of the expert is denoted by $V_{ij} \in \{1, \ldots , K\}$. 
It is notationally helpful to rewrite this vote into the $K$ dimensional indicator vector, which we denote in bold with $\boldsymbol{V}_{ij} = (\boldsymbol{1}\{V_{ij} = 1\},..., \boldsymbol{1}\{V_{ij} = K\})$, with $\boldsymbol{1}\{\cdot\}$ as indicator function. This allows to accumulate the annotators' votes into $\boldsymbol{Y}_i = (Y_{i1},..., Y_{iK})$ with $Y_{ik} = \sum_{j=1}^{J_i} \boldsymbol{1}(V_{ij} = k)$. This vector can be considered as the vote distribution for image $i$. To keep the notation simple we will drop index $i$ from the number of voters per image and write $J$ subsequently. We emphasize though, that images can be labeled by different numbers of voters, as our examples demonstrate. 

\subsection{Binary Case: $K=2$} 
For a more straightforward presentation and interpretation of our modeling strategy, we start with the binary case $K=2$. 
We assume a binary label representation, which is embedded into the two-dimensional space. That is, each instance (image or text)  is allocated with
\begin{align*}
    \boldsymbol{Z_i} = (Z_{i1},Z_{i2}) \in \mathbb{R}^2.
\end{align*} 
The vector $Z_i$ can be interpreted as {\bf embedding} or {\bf embedded ground truth values}, meaning that we represent the labeled instance $i$ as a point in a two-dimensional space. To simplify the notation we will drop the index $i$ in the following. 

\begin{figure}[h]
\centering
\includegraphics[width=0.4\textwidth]{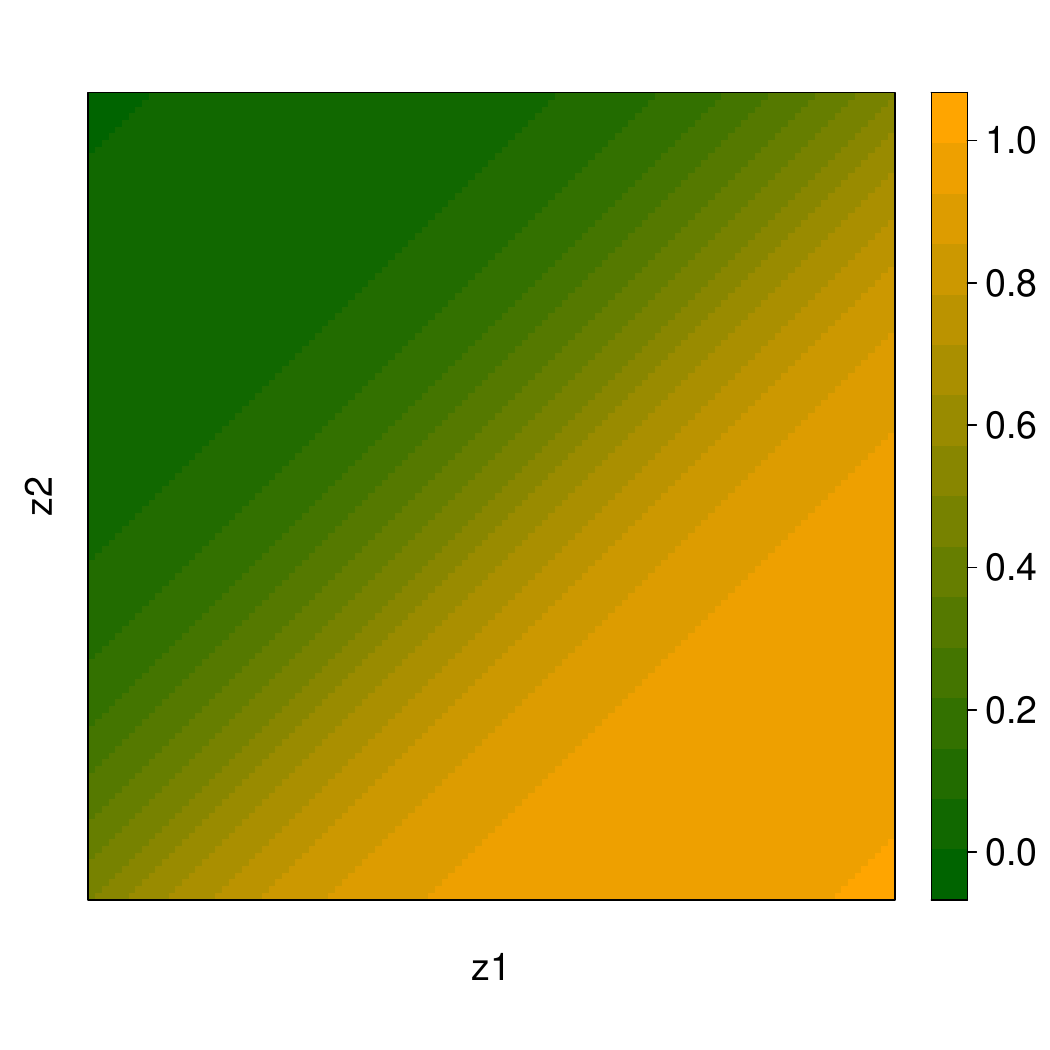}
\includegraphics[width=0.4\textwidth]{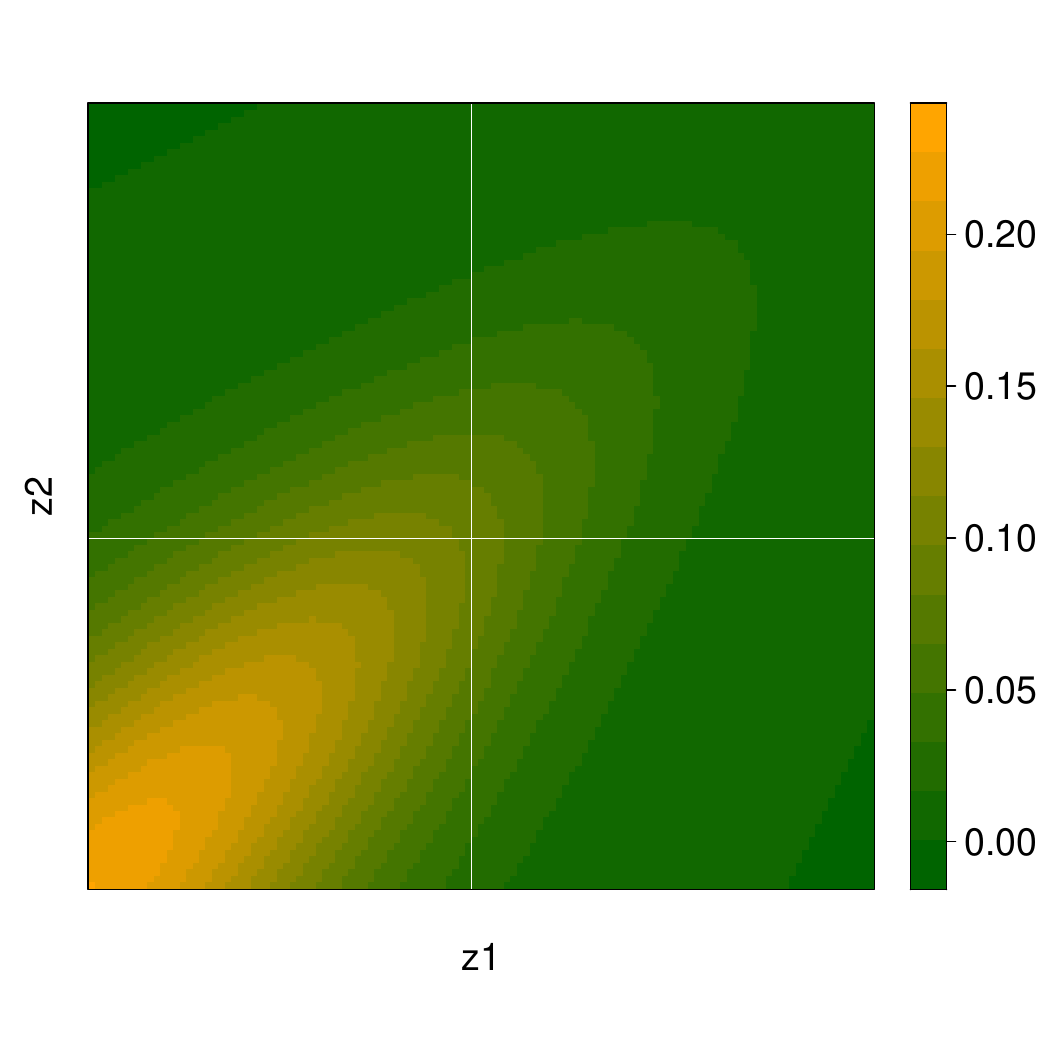}
\caption{\label{fig:meanbetabin} The figure shows the mean (left) and log-variance (right) of the Beta-Binomial distribution for different values of $\boldsymbol{Z}$, expressed through color.}
 \end{figure}

The embedding steers ambiguity as well as the uncertainty of the labeling process. This is achieved by relating $Z$ to the coefficients of a Beta distribution. To be specific we define $\alpha_Z=\exp(Z_1)$ and $\beta_Z=\exp(Z_2)$ as parameters of a Beta distribution, from which we draw the binomial parameter $\pi$ as 
\begin{align*}
    \pi \sim \mbox{Beta}(\alpha_Z, \beta_Z).
\end{align*} 
Given $\pi$, we obtain the image labels by drawing from the binomial distribution
\begin{align*}
    Y|\pi \sim B(J, \pi),
\end{align*} 
where $J$ is the number of votes or annotations of the respective image. Note that $J$ can vary for different instances, which for simplicity of notation we ignore here.
Apparently, if $\pi$ is close to 0 or 1, the image has no or little ambiguity, i.e.\ it is easy to label. 
Note that $\pi$ remains unobserved, so that given $Z$, we have 
\begin{align}
\nonumber
	P(Y=y|\boldsymbol{Z}) &\propto \int_{\pi} \binom{J}{y} \pi^y (1-\pi)^{(1-y)}\pi^{\alpha_Z} (1-\pi)^{\beta_Z} d\pi \\
	\label{eq:betabin}
	&\propto \binom{J}{y} \frac{B(\alpha_Z+y, \beta_Z+J-y)}{B(\alpha_Z,\beta_Z)},
\end{align}
where $B(.)$ denotes the univariate Beta function. \\
Within this model setup, we can derive a couple of interpretations. 
Interpreting $\boldsymbol{Z} \in R^2$ as ground truth,
we obtain the Beta-Binomial model (\ref{eq:betabin}). This in turn allows us to derive the mean value of $\pi$ through 
\begin{align*}
    E(\pi | \boldsymbol{Z}) = \frac{ \exp( Z_1 ) }{\exp(Z_1 ) + \exp(Z_2) }.
\end{align*}
Additionally, we can also quantify uncertainty by calculating the variance, which results as 
\begin{align*}
    Var(\pi | \boldsymbol{Z}) = \frac{ \exp( Z_1 ) \exp (Z_2)} {\big(\exp(Z_1 ) + \exp(Z_2)\big)^2 \big(\exp(Z_1 ) + \exp(Z_2) + 1 \big) }.
\end{align*}
For different values of $\boldsymbol{Z}$, we plot the mean and the (log)-variance of the Beta-Binomial distribution in Figure \ref{fig:meanbetabin}.
The variance expresses the uncertainty, which is how likely an image/text is misclassified given the data at hand. The larger $Z_1$, the more likely the instance is classified as ``one''. On the contrary, the larger $Z_2$, the more likely the image/text is classified as ``zero''. Moreover, the smaller the values of $Z_1 $ and $Z_2$ and the smaller the difference between them, the larger the variance. Hence, the concrete location of $Z_1$ and $Z_2$ expresses how likely it is that we can quantify an image or text in one category and how certain we are with respect to the class.

Apparently, $\boldsymbol{Z}$ is latent, but we aim to draw information about $Z$ given the votes $Y$. The two variables are connected via the Beta-Binomial model (\ref{eq:betabin}) and we can estimate the distribution of $Z$ for given votes $Y$ by drawing Markov Chain Monte Carlo (MCMC) samples.  
To do so, we sample from the posterior 
\begin{align*} 
f(\boldsymbol{Z}|Y=y) &\propto f(y|\boldsymbol{Z}) \cdot f_{prior}(\boldsymbol{Z}) \\ 
&\propto \binom{J}{y} \frac{B(\alpha_Z+y, \beta_Z+J-y)}{B(\alpha_Z,\beta_Z)}  \cdot f_{prior}(\boldsymbol{Z}).
\end{align*}
As prior distribution for $\boldsymbol{Z}$, we use a bivariate normal with mean $\boldsymbol{\mu}$ and variance matrix $\boldsymbol{\Sigma}$ and estimate these parameters following an empirical Bayes approach. 
While this prior distribution is presumably too simple to completely capture the true hidden structure in the data and hence does not constitute the underlying data-generating process, it is a convenient modeling assumption. Postulating a multivariate Gaussian distribution for the latent embedded ground truth provides numerical stability during estimation and allows to express uncertainty, even if only one annotation is available. 
%

With $Y_i$ as votes on image $i$, $J_i$ referring to the number of annotations on image $i$ and $\boldsymbol{Z}_i \in \mathbb{R}^2$ as the embedded ``true location'' of image $i$ within the two-dimensional space, we run the estimation algorithm laid out in Figure \ref{fig:mcmc_embedding}. We also refer to  Appendix \ref{sec:appendix-algorithm} for additional computational details. As a result, we obtain estimates $\hat{\boldsymbol{z}}_i$ for each image by averaging the MCMC samples for instance $i$ of the last iteration.
The point itself expresses the classification difficulty for the respective image and is thus far more informative than a singular ground truth label.

\begin{figure}[ht]
\par\noindent\rule{\textwidth}{0.4pt}
  \begin{enumerate}
    \item Initialize the prior parameters $\boldsymbol{\mu}$ and $\boldsymbol{\Sigma}$ and set them to $\boldsymbol{\mu}^{(m)}$ and $\boldsymbol{\Sigma}^{(m)}$, $m=1$. 
    \item For $i = 1, \ldots, n$: Draw an MCMC sample from 
            \begin{align*}
        f(\boldsymbol{Z}_i|Y_i=y_i) &\propto \binom{J_i}{y_i} \frac{B(\exp(Z_{i1})+y_i, \exp(Z_{i2})+J_i-y_i)}{B(\exp(Z_{i1}),\exp(Z_{i2}))} \\
        &\cdot \exp(-0.5(\boldsymbol{Z}_i-\boldsymbol{\mu}^{(m)})^T \cdot \boldsymbol{\Sigma}^{(m)-1} \cdot (\boldsymbol{Z}_i-\boldsymbol{\mu}^{(m)}))
            \end{align*}
	
	and take $\hat{\boldsymbol{z}}_i^{(m)} = E(\boldsymbol{Z}_i|Y_i)$ as the mean of the MCMC samples.
    \item Update 
	\begin{align*}
		&\hat{\boldsymbol{\mu}}^{(m+1)} = \frac{1}{n} \sum_{i=1}^n \hat{\boldsymbol{z}}_i^{(m)} \\
		& \hat{\boldsymbol{\Sigma}}^{(m+1)} = \frac{1}{n} \sum_{i=1}^n (\hat{\boldsymbol{z}}_i^{(m)} - \hat{\boldsymbol{\mu}}^{(m+1)})(\hat{\boldsymbol{z}}_i^{(m)} - \hat{\boldsymbol{\mu}}^{(m+1)})^T
	\end{align*}
	and repeat from Step 2.
  \end{enumerate}
\par\noindent\rule{\textwidth}{0.4pt}
  \caption{Estimating the embeddings.}\label{fig:mcmc_embedding}
\end{figure}

\subsection{Multiclass Case: $K>2$} 

The binary model can now be easily extended to more than two classes by employing the Dirichlet distribution. Now, $\boldsymbol{Z}=(Z_1,...,Z_K) \in \mathbb{R}^K$ is the embedded ground truth and we obtain the parameters 
$ \alpha_k = \exp(Z_k), \forall k=1,...,K$. From these, we can draw 
\begin{align*}
    \boldsymbol{\pi} \sim Dir(\alpha_1,...,\alpha_k). 
\end{align*}
This results in the multinomial parameter vector $\boldsymbol{\pi} = (\pi_1,...,\pi_K),$ where $K$ corresponds to the number of classes. 
Given $\boldsymbol{\pi}$, the votes are assumed to be drawn from a multinomial distribution:
\begin{align*}
    \boldsymbol{Y}|\boldsymbol{\pi} \sim Mult(\boldsymbol{\pi}, J),
\end{align*} 
with $J$ denoting the number of votes. This leads to the two probability functions
\begin{align*}
&f(\boldsymbol{Y}=\boldsymbol{y}|\boldsymbol{\pi}) = \frac{J!}{y_1!... y_K!} \prod_k \pi_k^{y_k} \\
&f(\boldsymbol{\pi}|\boldsymbol{\alpha}) = \frac{1}{B(\boldsymbol{\alpha})} \prod_k \pi_k^{\alpha_k-1}.
\end{align*}
In this case, the function $B(.)$ denotes the multivariate version of the Beta function. 
The vector $\boldsymbol{\pi}$ remains unobserved and we can calculate the probability of $\boldsymbol{Y}$ given $\boldsymbol{Z}$ by marginalizing over $\boldsymbol{\pi}$:
\begin{align*}
P(\boldsymbol{Y}=\boldsymbol{y}|\boldsymbol{\alpha}) 
&= \int_\theta f(\boldsymbol{y}|\boldsymbol{\pi}) f(\boldsymbol{\pi}|\boldsymbol{\alpha}) d \boldsymbol{\pi} = \frac{J!}{y_1!... y_K!} \cdot \frac{1}{B(\boldsymbol{\alpha})} \cdot \int_{\boldsymbol{\pi}} \prod_k \pi_k^{y_k+\alpha_k-1} d\boldsymbol{\pi} \\
&= \frac{J!}{y_1!... y_K!} \cdot \frac{\Gamma(\sum \alpha_k)}{\prod_k \Gamma(\alpha_k)} \cdot \frac{\prod_k (\Gamma(\alpha_k+y_k))}{\Gamma (\sum_k \alpha_k+y_k)} = \frac{J!}{y_1!... y_K!} \cdot \frac{\Gamma(\sum \alpha_k)}{\Gamma (\sum_k \alpha_k+y_k)} \cdot \prod_k \frac{\Gamma(\alpha_k+y_k)}{\Gamma(\alpha_k)} \\
&= \frac{J!}{y_1!... y_K!} \cdot \frac{B(\boldsymbol{\alpha} +\boldsymbol{y}) }{B(\boldsymbol{\alpha})}.
\end{align*}

Again, the embedded ground truth values $\boldsymbol{Z}$ can be estimated given the votes $\boldsymbol{Y}$ using MCMC samples with the stochastic EM algorithm. Following the binary case and assuming a multivariate Gaussian prior for the embeddings $\boldsymbol{Z}$ leads to the posterior distribution
\begin{align*}
f(\boldsymbol{Z}|\boldsymbol{Y}) \propto f(\boldsymbol{Y}|\boldsymbol{Z}) f_{\text{prior}}(\boldsymbol{Z}).
\end{align*} 

We obtain a Dirichlet-Multinomial model by assuming a $K$-dimensional embedded ground truth $\boldsymbol{Z}$ for each image. The parameter $\boldsymbol{\pi} =(\pi_1,...,\pi_K)$ follows a Dirichlet distribution given $\boldsymbol{Z}$, and we can easily derive expectation and variance for all entries $Z_{k} \in \boldsymbol{Z}, k=1,...,K$: 
\begin{align*}
    E(\pi_{k}|\boldsymbol{Z}) &= \frac{\exp(Z_{k})}{\sum_{k'=1}^K \exp(Z_{k'})} \\
    Cov(\pi_{k}, \pi_{k'}|\boldsymbol{Z}) &= \frac{1}{1+\sum_{k'=1}^K exp(Z_{k'})} \cdot \frac{\exp(Z_{k})}{\sum_{k'=1}^K \exp(Z_{k'})} \cdot \left(1-\frac{\exp(Z_{k})}{\sum_{k'=1}^K \exp(Z_{k'})}\right).
\end{align*}

Each entry of $\boldsymbol{Z}$ corresponds to one of the $K$ classes. The concrete values can again be interpreted in two ways. On the one hand, $Z_k$ hints at how likely the image is classified into category $k$. On the other hand, the difference between the entries of $\boldsymbol{Z}$, i.e. the distance between classes $k$ and $k'$ for $k' \neq k$, corresponds to the certainty about the category $k$ versus $k'$.

Following the estimation procedure described in Section 2.2 adapted to the multiclass case leads to values $\hat{\boldsymbol{z}}_i$ for all images $i=1,...,n$.  Like above, these values form an {\bf embedding} of the image in the $K$ dimensional space. 
These embeddings express the classification (un-)certainty of the individual images in terms of distribution. The concrete algorithm is comparable to the case $K=2$ discussed above and therefore not explicitly laid out here again. 
\section{Results}

To showcase the generality and versatility of the proposed approach in various applications, the proposed model is applied to the three datasets described in the introduction.
The datasets are typical examples in the field of multiple annotations and annotator disagreement and hence provide ground for analyzing the uncertainty associated with the labels. 
Table \ref{tab:datasets} contains general information about the three datasets discussed in this section. 

\begin{table}
\centering
\begin{tabular}{c|cccc}
	Dataset & \#Images $N$ & \#Classes $K$ & \#Distinct Annotation Patterns & \#Annotations $J$
	 \\ \hline 
  ChaosSNLI & 1514 & 3 & 832 & 100 \\
So2Sat LCZ42 & 159581 & 16 & 360 & 11 \\
Cifar-10H & 10000 & 10 & 3406 & [50,63] 
\end{tabular}
\vspace{5pt}
\caption{Overview of the datasets.}
\label{tab:datasets}

\end{table}

\subsection{ChaosSNLI}

First, we explore the advantages of the proposed methodology in the context of the classification of language, i.e.\ the domain of NLI, as shortly introduced in Section \ref{sec:intro}.
The multi-annotator dataset ChaosSNLI\footnote{Download available from \url{https://github.com/easonnie/ChaosNLI} (accessed on 04th of February 2024)} is based on the development set of the Standford Natural Language Inference (SNLI) dataset (\citealp{bowman:2015}) and was introduced in the context of label ambiguity by \cite{nie:2020}. 
It contains multiple annotations for sentence pairs, i.e.\ pairs of premise and hypothesis. For each premise, three hypotheses are originally generated by an annotator, as an entailing, neutral, and contradicting description of the premise. The resulting sentence pairs of premise and hypothesis can therefore be classified as \textit{entailment, neutral} or \textit{contradiction}. Note that a subjective ground truth, namely the original intention of the first annotator, is available for this specific dataset. However, it cannot be recovered by the annotators in many cases and the dataset ChaosSNLI especially showcases this problem as it contains sentence pairs exhibiting a high rate of disagreement. Particularly, $N=1514$ sentence pairs are re-assessed by a large number of annotators, i.e. $J=100$, and assigned to one of the three classes, as shown in Table \ref{tab:intro_examples}. 
Due to the ambiguous nature of language and the individual perception, the disagreement rate in the annotations is high and the original true label cannot be recovered reliably. 
For the classification of language, the existence of a single ``gold'' label is especially doubtful and hence, the need for alternative and more appropriate representations of labels persists. 
Applying the methodology proposed in Section \ref{sec:model} allows us to estimate embedded ground truth vectors for the observations based on the provided annotations, which will be analyzed in this subsection. 

\begin{figure}
    \centering
    \begin{tabular}{lll}
    \footnotesize
    \textbf{ID} & 
        \textbf{Context/Premise} & \textbf{Statement/Hypothesis} \\ \hline 
          34 & A man running a marathon talks to his friend. & There is a man running. \\
         1168 & A black and white dog running through shallow water. &  Two dogs running through water. \\
         1177 & A woman holding a child in a purple shirt. & The woman is asleep at home. \\
         1371 & An elderly woman crafts a design on a loom. & The woman is sewing. \\
         &&
    \end{tabular}
    \includegraphics[width=0.45\textwidth]{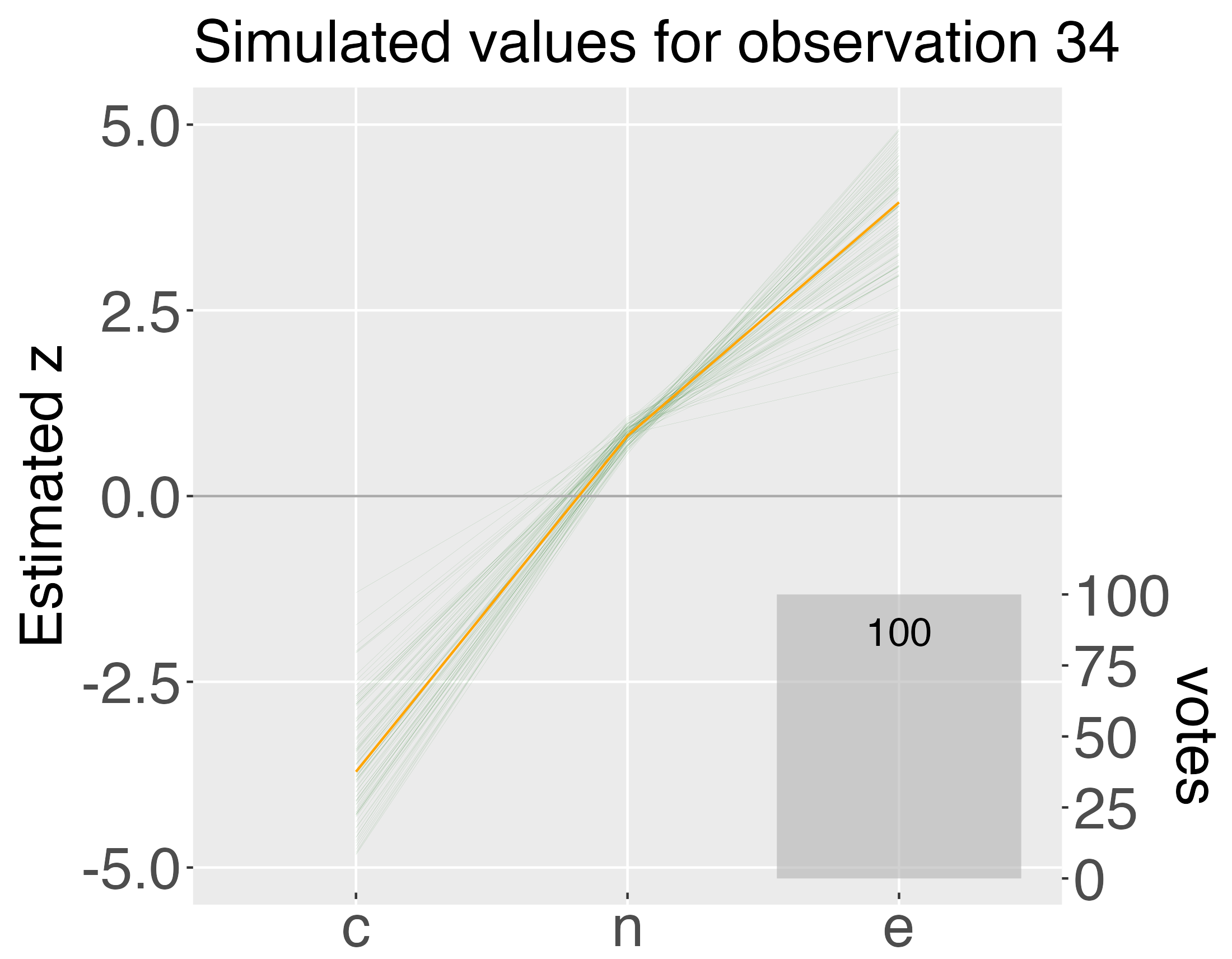}
    \includegraphics[width=0.45\textwidth]{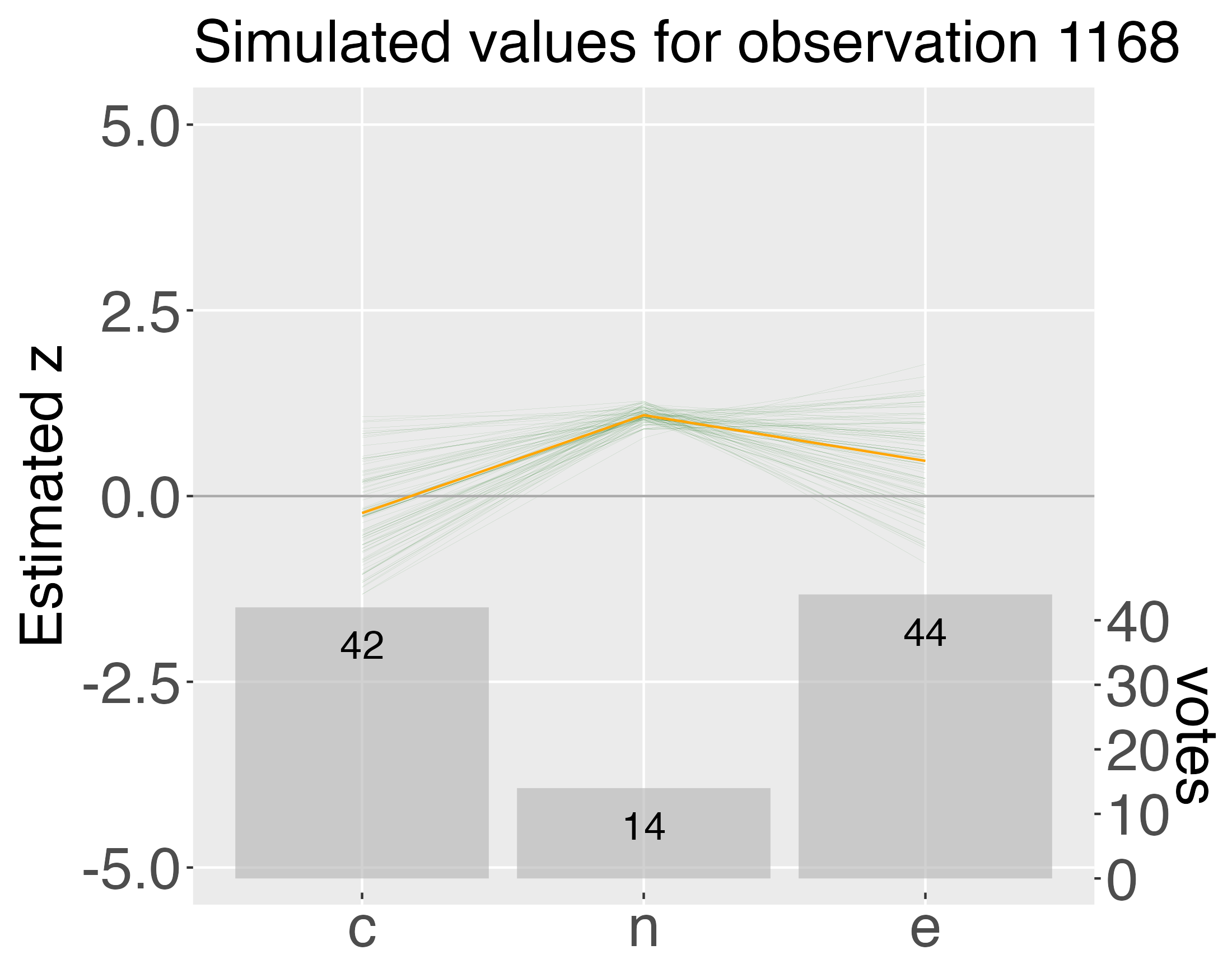}
    \includegraphics[width=0.45\textwidth]{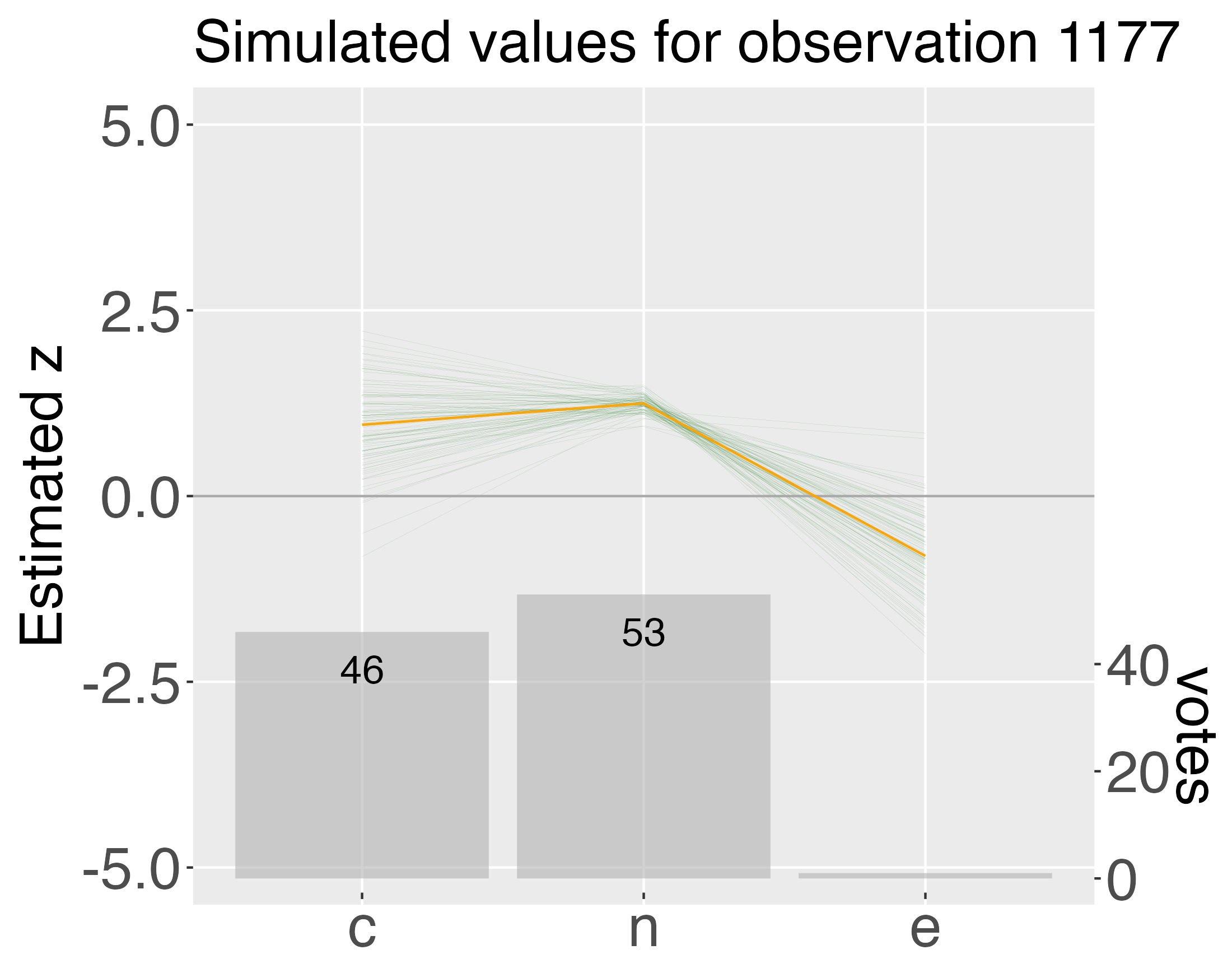}
    \includegraphics[width=0.45\textwidth]{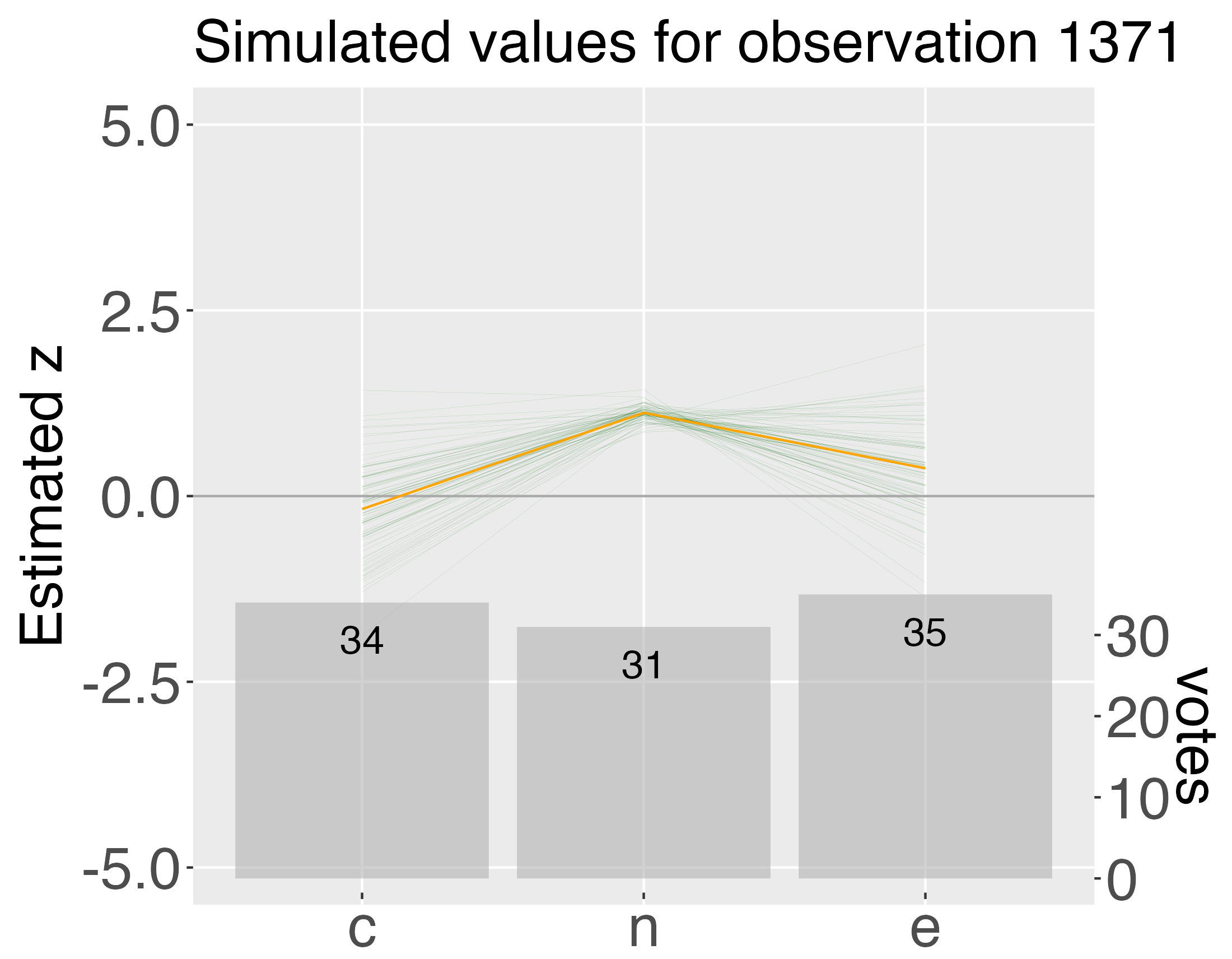}
    \caption{The plots show the estimated embedded ground truth vectors for exemplary sentence pairs from the dataset ChaosSNLI. The actual estimated vector is shown as orange line, the green lines represent the MCMC samples and the actual annotations are shown as grey bars.}
    \label{fig:chaosnli_embedding_examples}
\end{figure}

First, let us return to the \textbf{exemplary sentence pairs} provided in Table \ref{tab:intro_examples} and inspect the respective embeddings. Figure \ref{fig:chaosnli_embedding_examples} shows the estimated values for the exemplary sentence pairs, along with the observed annotations as well as the MCMC samples. While the estimated values for observation 34 (upper left plot) express clear affiliation to class \textsl{entailment}, all other embeddings reflect the ambiguity within the sentence pairs and also the associated annotations. Not only are the class-specific estimated values rather small and hence similar across the three categories, we also see a tendency towards class \textsl{neutral} for ambiguous instances, even though the majority vote might advocate otherwise. This expresses ambiguity with respect to classification and the ``weaker'' interpretation of class \textsl{neutral} compared to the other two categories. If all three classes received annotations, it is semantically unlikely that the respective instance can be uniquely classified into either \textsl{entailment} or \textsl{contradiction}. Hence, the model favours class \textsl{neutral} in such situations.

\begin{figure}
    \centering
    \includegraphics[width=0.7\textwidth]{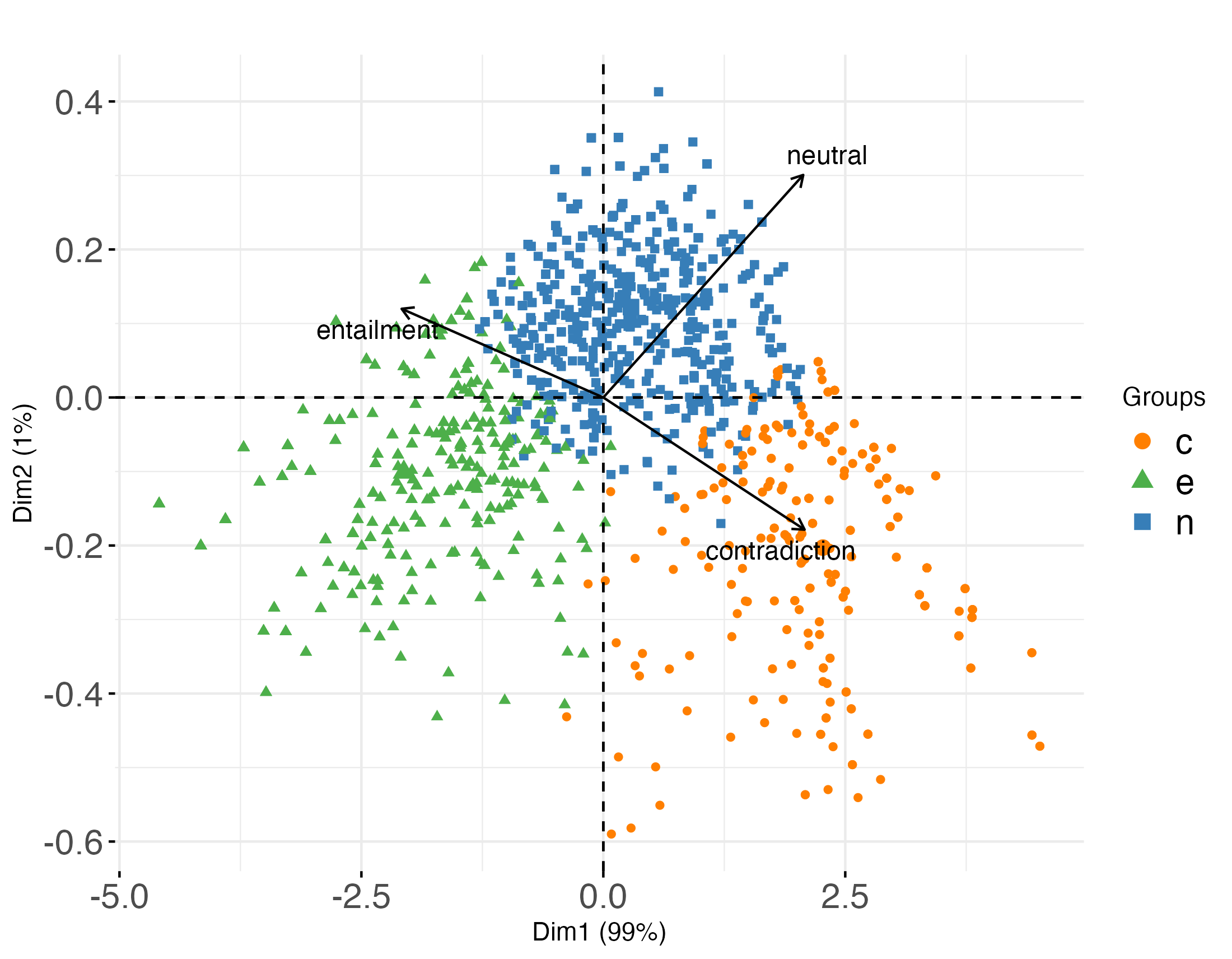}
    \caption{The biplot shows the estimated embeddings for ChaosSNLI, projected into two dimensions. The scatterpoints represent the instances, colored by majority vote. The original dimensions are represented as arrows.}
    \label{fig:biplot_chaosnli}
\end{figure}

For this particular application, the classes themselves are by definition uncorrelated or negatively correlated. This property is also expressed by the resulting estimated embeddings. To visually inspect the results, we employ dimensionality reduction techniques for easier exploration. We specifically utilize principal component analysis (PCA) to extract the principal components from the estimated embeddings. PCA is a widely used technique for linear dimensionality reduction, commonly employed for exploratory analysis and visualization of high-dimensional data. For detailed information on the methodology, refer to e.g.\ \cite{jolliffe:2002}.
Here, we especially focus on the visualization benefits of the technique. Namely, it is possible to plot the observations in a so-called two-dimensional \textbf{biplot} after applying PCA. Figure \ref{fig:biplot_chaosnli} shows the respective plot for the dataset ChaosSNLI. The estimated embeddings are projected onto the two-dimensional space, spanned by the two first principal components. The embeddings of the instances are visualized as scattered points, where their overall similarity is expressed by proximity. In this case, no specific clustering is apparent. By coloring the scattered points according to the observed majority voting, we observe some overlap of the singular classes in the two-dimensional embedding space. 
Additionally, the vectors correspond to the original variables, i.e. the categories and dimensions of the embeddings. The angles between the vectors express the degree of correlation between the variables, i.e. small angles suggest a high positive correlation. However, in this case, the angles between variable \textsl{neutral} and the other two variables are roughly 90 degrees, indicating no correlation between the quantities. In contrast, the angle between variables \textsl{entailment} and \textsl{contradiction} is close to 180 degrees, hence expressing a negative correlation. Of course, this is reasonable from a semantic perspective and in line with the interpretation of the classes. Additional results for the ChaosSNLI dataset are reported in Appendices \ref{sec:appendix:m-vote} and \ref{sec:appendix-annotations}.

\subsection{So2Sat LCZ42}
To showcase the generality of the proposed approach, we now move on to the analysis of a multi-annotator dataset for image classification.
In the domain of image classification, the assumption of a singular ground truth label is especially doubtful if the images themselves are ambiguous. A prominent example is the classification of low-resolution satellite images.
We analyze the earth observation benchmark dataset So2Sat LCZ42\footnote{Download available from: \texttt{https://mediatum.ub.tum.de/1659039} (accessed on 03rd of Febuary 2024)}, see Table \ref{tab:datasets} for a brief overview. For a detailed description of the complete dataset, we refer to \citet{Zhu:2020}. The task is to classify satellite images into one of 17 Local Climate Zones (LCZ). See also Figure \ref{fig:intro_examples} for a sketch of the LCZ. Note that zone 1 to 10 refer to urban areas while zones A to G are non-urban.  The data comprises humanly classified satellite images from urban agglomerations in Europe, of which some images were assessed multiple times during the labeling phase to ensure label quality. Overall, we look at images from 9 European cities with multiple annotations. Figure \ref{fig:intro_lcz_scheme} shows the categories for the classification of satellite images. 
The distribution of votes across the classes is quite imbalanced, and particularly LCZ 7  rarely occurs in the dataset. Due to its composition (lightweight low-rise building types, i.e. ``slums''), it is doubtful to be observed in European cities. Therefore, we excluded LCZ 7 and restricted the model to $K=16$ instead of $K=17$, see also \cite{hechinger:2024}.
The voting patterns suggest that most images seemed easy to classify. For 77.18 \% of the images, the annotators agree on one single LCZ. 
Again, we apply the proposed framework and estimate an embedded ground truth value $\hat{z}^{(i)}$ for each instance represented as a vector of annotations.

\begin{figure}[h]
    \centering
    \includegraphics[width=0.49\textwidth]{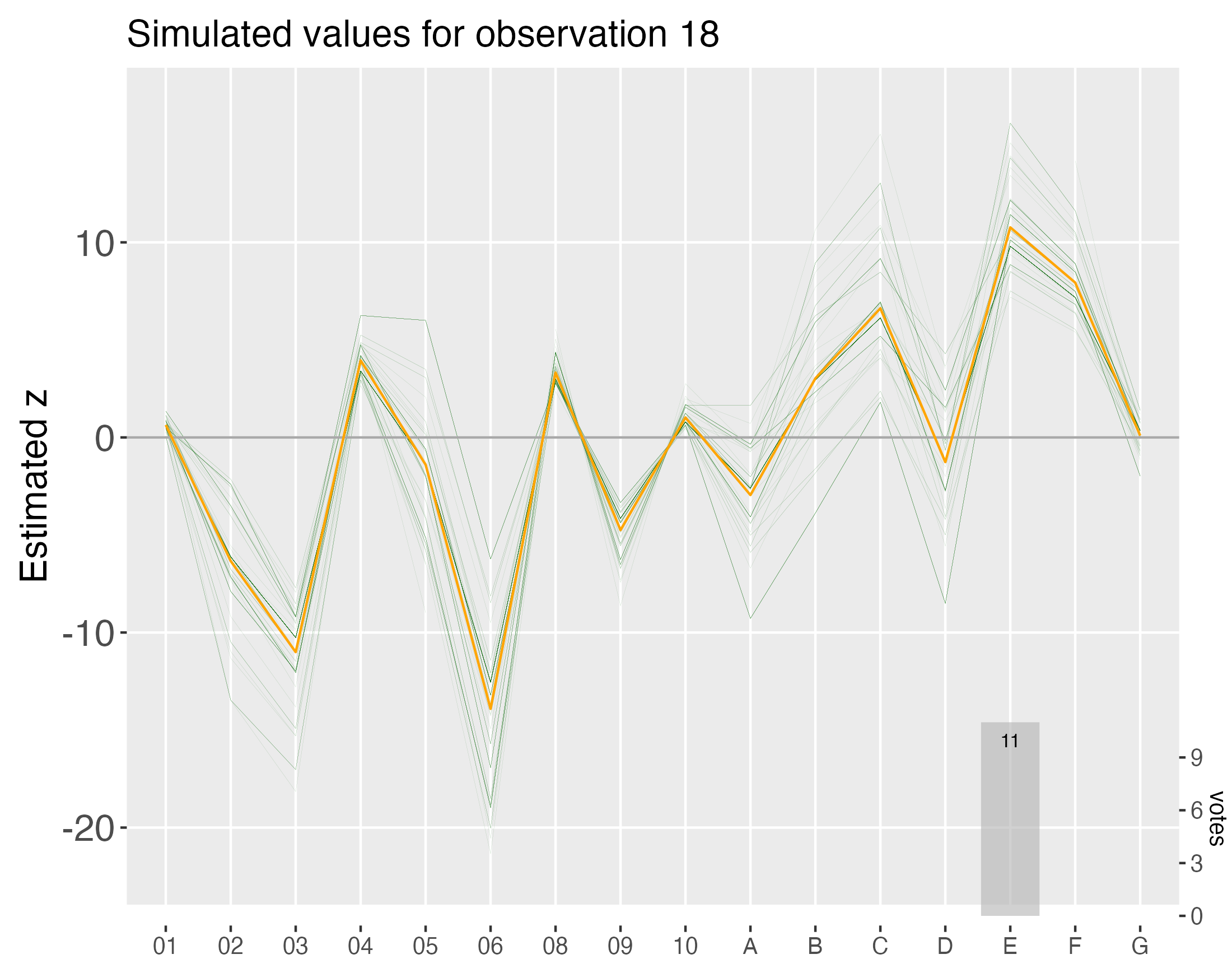}
    \includegraphics[width=0.49\textwidth]{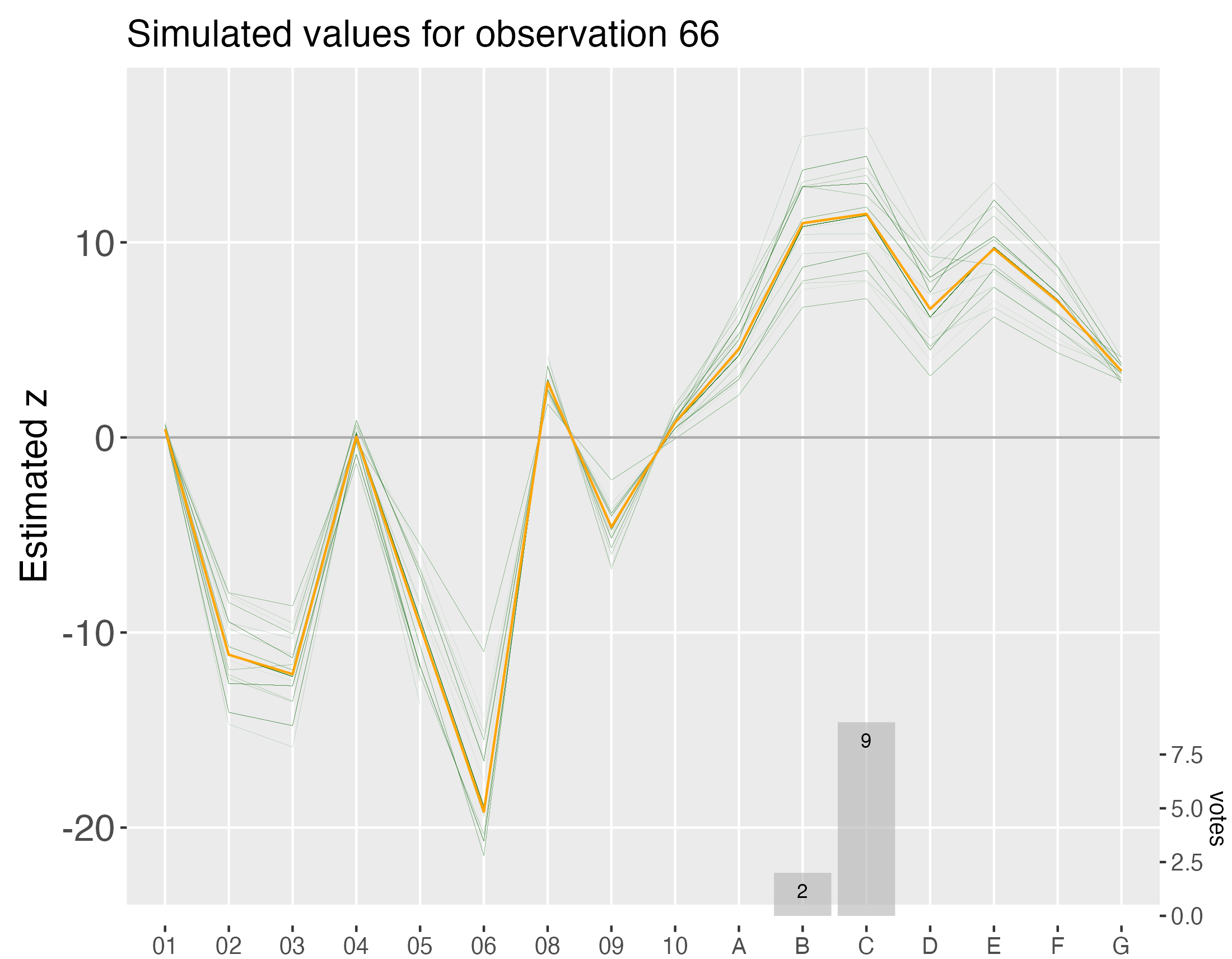}
    \includegraphics[width=0.49\textwidth]{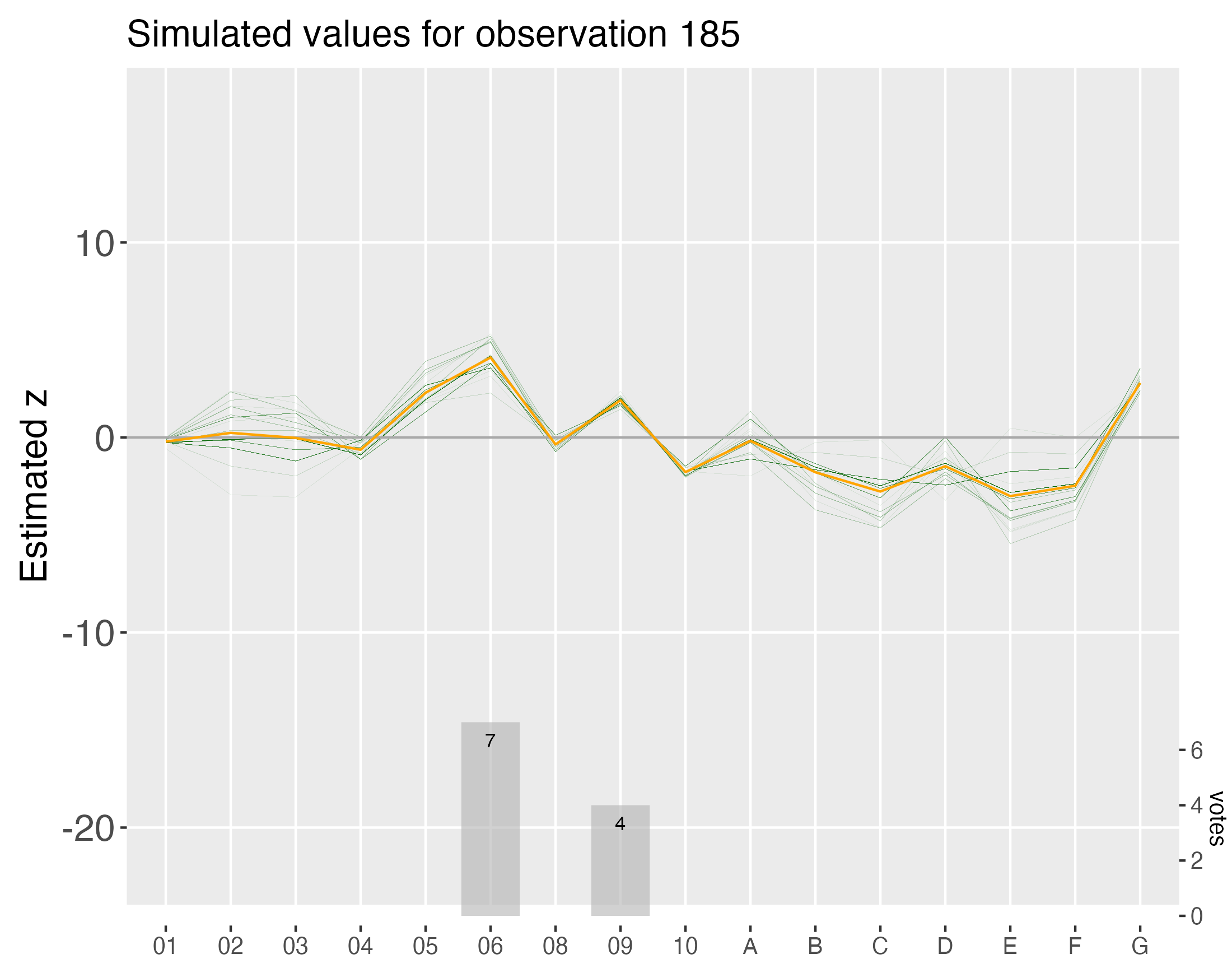}
    \includegraphics[width=0.49\textwidth]{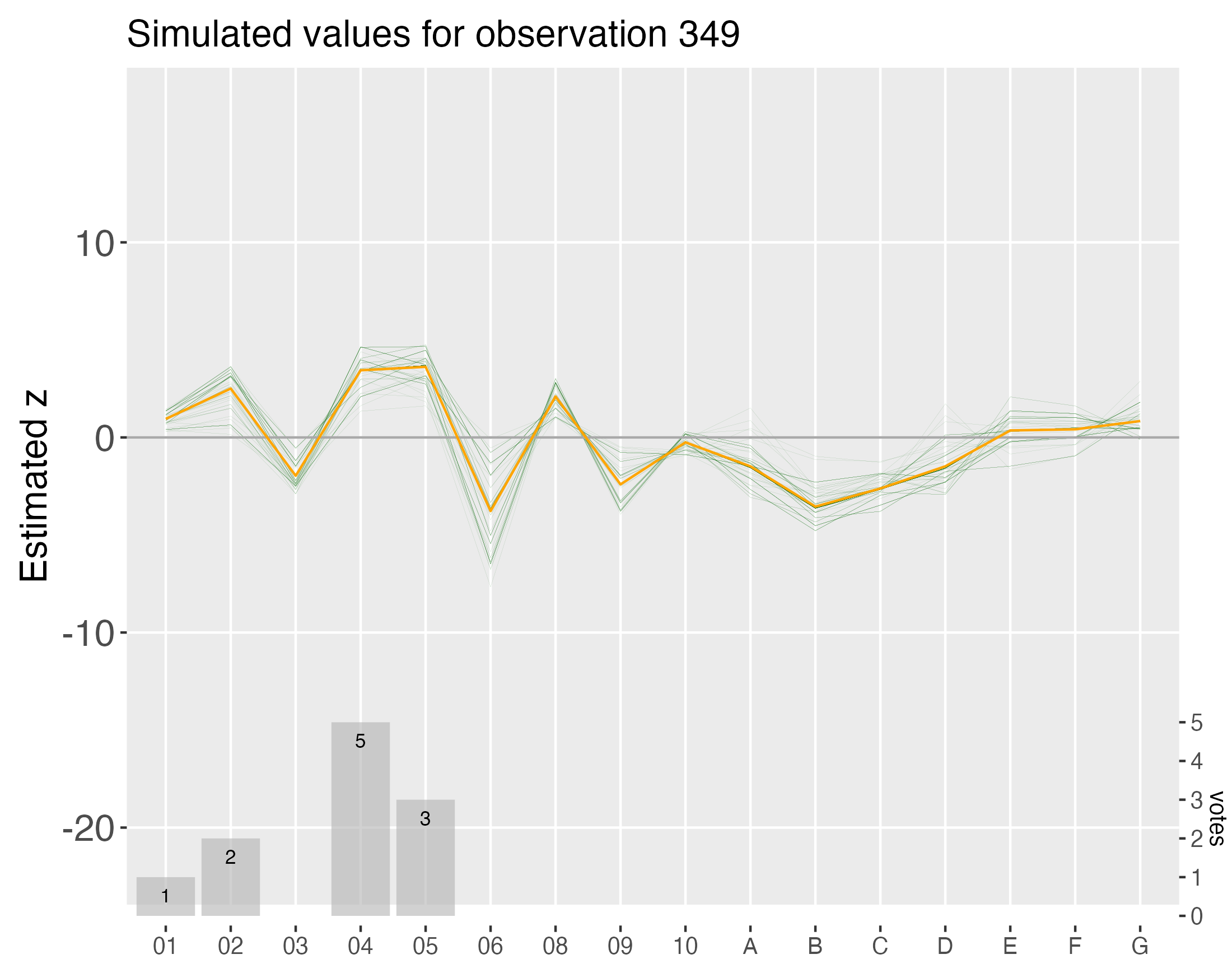}
    \caption{The plots show the MCMC samples as green lines and their mean value as an orange line for exemplary randomly chosen images. Additionally, the actual votings from the experts are shown as grey bars.}
    \label{fig:so2sat_embedding_examples}
\end{figure}

First, it is again helpful to examine the explicit estimates for some \textbf{exemplary instances}. 
We get the sampled embeddings from Step 2 in the algorithm shown in Figure \ref{fig:mcmc_embedding}, where the parameters are set to the final estimated parameters.
Figure \ref{fig:so2sat_embedding_examples} shows the explicit estimates, as well as the MCMC samples for four random images. The resulting posterior mean value (i.e.\ the estimated ground truth) is shown as an orange line and the observed votes as grey bars at the bottom of the plot. 
The first image with ID 18 (upper left plot) shows a non-confusion case, i.e.\ all experts agreed on class E. Looking at the ground truth estimate, we see a clear spike for the respective class, indicating that classification for this image is rather easy. 
Image 66 (upper right plot) gets nine votes for class C and two votes for class B, i.e. the voting pattern is restricted to non-urban classes. Here, we observe negative values for urban classes 1-10 and small positive values for all non-urban classes. Hence, the voting pattern is reflected in the estimated ground truth in terms of the magnitude of the values.
Image 185 (lower left plot) shows confusion of two urban classes, namely classes 6 and 9. In this case, the estimated ground truth values clearly correspond to these votes with almost all other values being negative or very close to zero. The image might be ambiguous regarding the correct ground truth, as those two values are quite close. 
Lastly, confusion happened for urban classes only for the image with ID 349 (lower right plot). The majority of experts voted for class 4, three experts chose class 5, class 2 received 2 votes and one expert voted for class 1. This is also reflected in the estimated $\hat{z}_k$ values. We estimate similar positive values for classes 4, 5 and 2 and a slightly smaller value for class 1, reflecting the voting patterns but also incorporating additional information about the confusion risk of the classes.

\begin{figure}
    \centering
\begin{subfigure}{0.49\textwidth}
    \includegraphics[width=\textwidth]{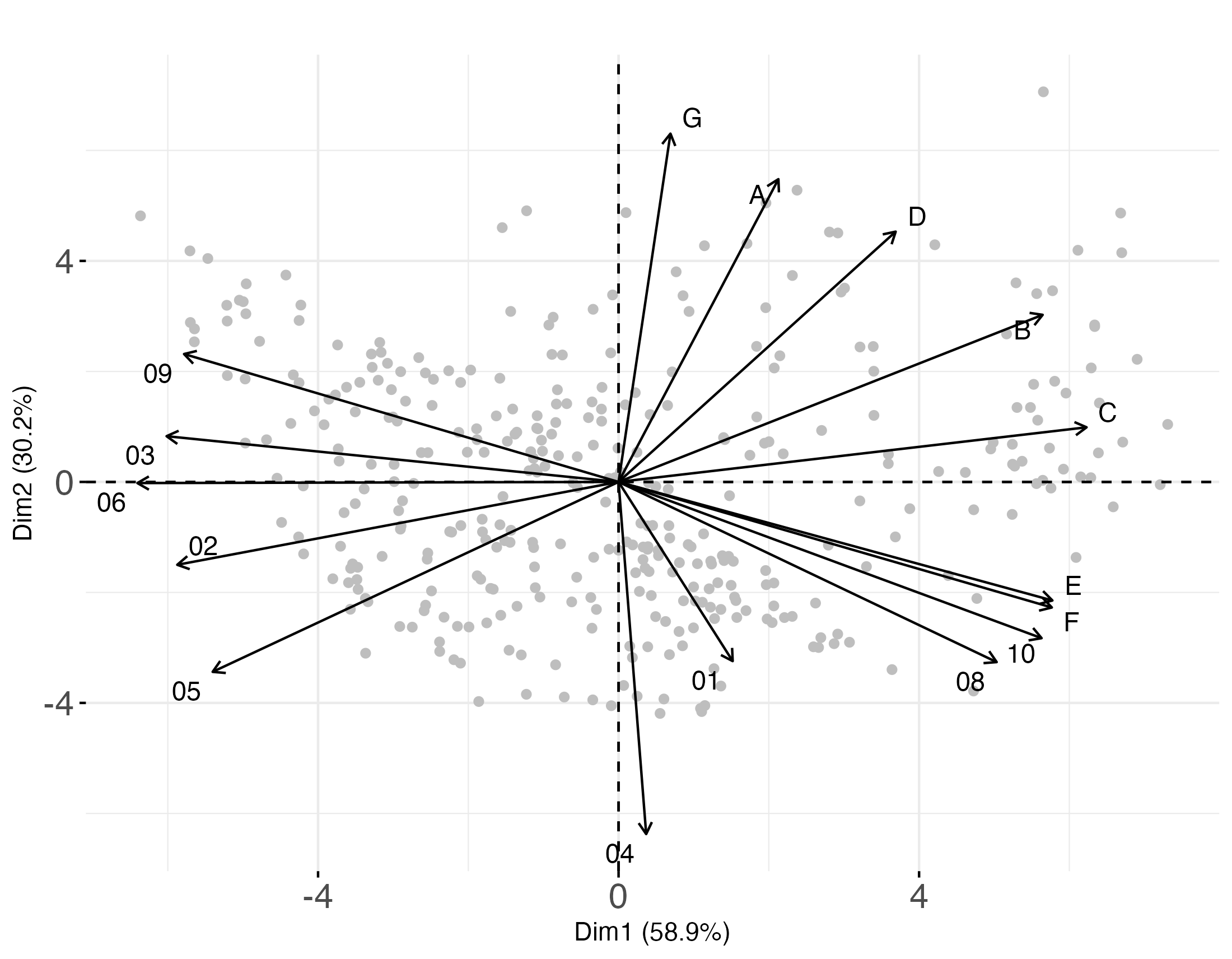}
    \caption{Biplot.}
    \label{fig:biplot_so2sat}
\end{subfigure}
\begin{subfigure}{0.49\textwidth}
    \includegraphics[width=\textwidth]{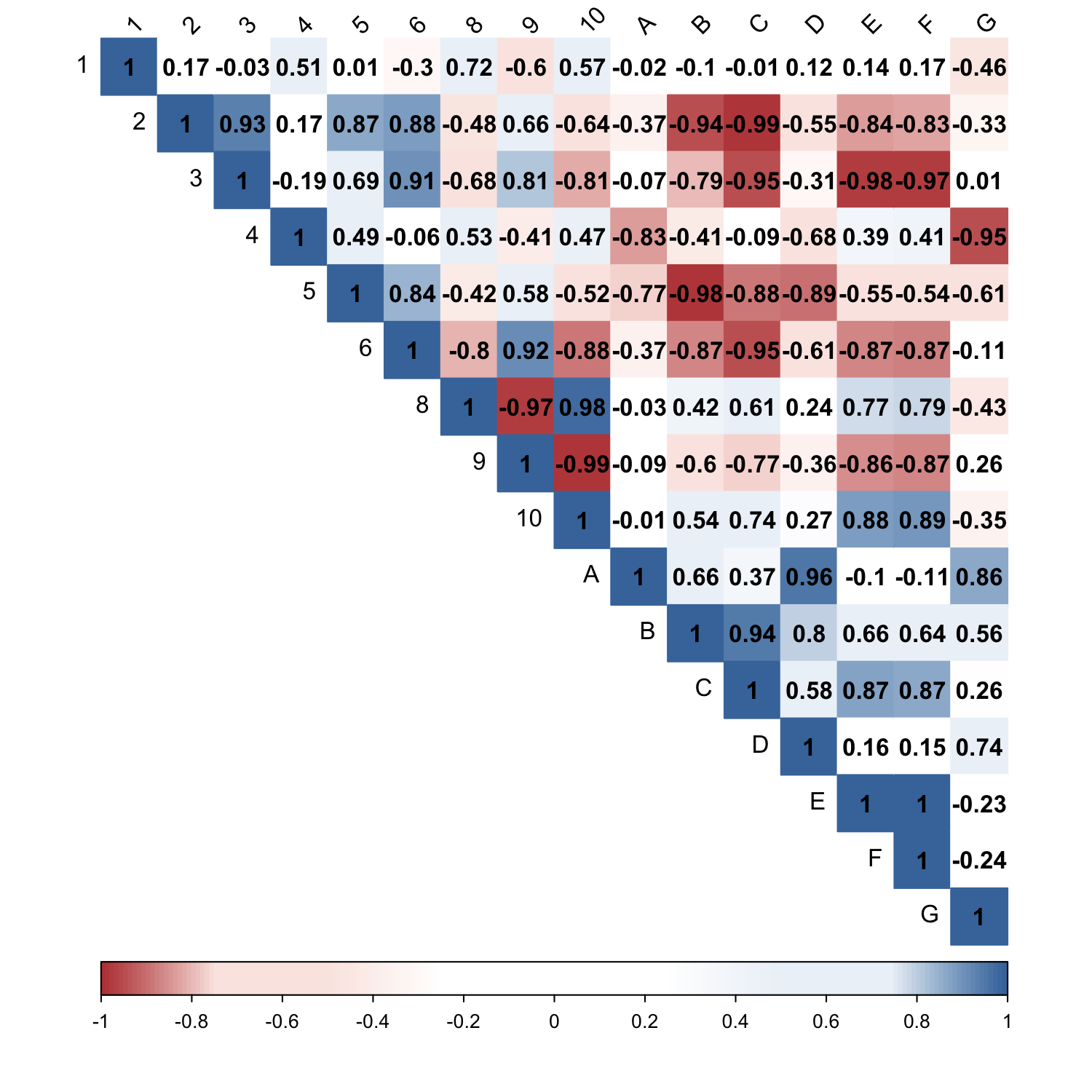}
    \caption{Correlation Matrix.}
    \label{fig:corr_so2sat}
\end{subfigure}
\caption{The subfigures show additional results for the dataset So2Sat-LCZ42. The biplot of the projected embeddings, as well as the correlation matrix of the estimates, show similarities of the categories leading to ambiguities and confusion.}
\label{fig:so2sat}
\end{figure}

While low-resolution satellite images are of course prone to ambiguity, in this particular application, the classes themselves are also ambiguous and not easy to distinguish (\citealp{hechinger:2024}). 
Hence, the ambiguity of the classes contributes to the uncertainty in the annotations. This property is nicely expressed by the vectors shown in the \textbf{biplot} in Figure \ref{fig:biplot_so2sat}, as well as the \textbf{estimated correlation matrix} of the ground truth embeddings in Figure \ref{fig:corr_so2sat}.
This matrix can be interpreted as a generalized confusion matrix, in the case where a ground truth is not guaranteed. A high positive correlation indicates a high confusion risk. On the contrary, negative correlation values suggest that classes are well distinguishable.
The correlation patterns between classes are visible in the correlation matrix, see Figure \ref{fig:corr_so2sat}. 
Exemplary, we investigate the correlation values for LCZ 2. Apparently, it is positively correlated with classes 3, 5 and 6, shows weaker positive correlations to the other urban classes 1, 4 and 9 and is negatively correlated to all other classes. A semantic inspection of the description of the classes, given in Figure \ref{fig:intro_lcz_scheme}, makes this observation rather plausible. Class 2 refers to compact middle-rise areas and can therefore easily be confused with either other compact (LCZ 3), mid-rise areas (LCZ 5) or also open low-rise areas (LCZ 6), while distinction from classes 1, 4 and 10 as well as from nonurban classes A-G should be less difficult. 
We also see a strong correlation between class 8 and class 10, i.e. confusion of large low-rise areas (8) and heavy industry areas (10). This is again reasonable from a semantic perspective. 
Concerning nonurban classes A-G, we observe a higher correlation between classes A (dense trees) and D (low plants) as well as between B (scattered trees) and C (bush/scrub). The same holds for classes E and F, referring to bare rock/paved areas and bare soil/sand. Again, these classes are similar in their composition and confusion or overlap might occur. 
The matrix also shows that urban and nonurban classes (LCZ 1-10 vs. LCZ A-G) have mostly negative or close to zero correlation values, supporting the claim that confusion about these two types of LCZs rarely occurs.
Overall, the entries of the estimated ground truth estimate are correlated in a way that has a meaningful interpretation and aligns with the initial hypotheses. 
The MCMC procedure additionally enables us to inspect the variation of the correlation matrix. The variances of the individual values are quite small and on average very close to zero, indicating a stable estimation and reliable results. The standard deviations of the entries in Figure \ref{fig:corr_so2sat} are displayed in Appendix \ref{sec:appendix-std}. 

\subsection{Cifar-10H} 
The third dataset is a version of Cifar-10, a popular benchmark dataset for image classification, as introduced by \cite{krizhevsky:2009}. The subset Cifar-10H\footnote{Download available from \url{https://github.com/jcpeterson/cifar-10h} (accessed on 24th of August 2023)} as introduced by \cite{peterson:2019} contains multiple annotations for images in the test set, reflecting the uncertainty stemming from differences in human perception. Here, the natural images are categorized into unambiguous classes, see Table \ref{tab:datasets}. The original dataset has been extended with soft labels, i.e.\ multiple annotations, to achieve better generalization for classification models, specifically on out-of-sample datasets, see \cite{peterson:2019} and \cite{battleday:2020}. Therefore, $N=10000$ images of $K=10$ classes from the test set of Cifar-10 were annotated by 2571 Amazon Mechanical Turk workers. After an initial training phase, each worker annotated 200 images, 20 per category. To identify and remove low-performance annotators, attention checks were introduced after every 20 trials. \\
The current setting differs from the previously discussed dataset. Most images belong to one of the unambiguous categories and can be reliably classified by untrained annotators. Nevertheless, it is helpful to additionally inspect label embeddings reflecting the individual human perception, which can still be ambiguous. Due to the small size of the images, the pictured class is also not always identifiable, as shown in Figure \ref{fig:intro_examples}. The dataset contains a high degree of human consensus due to its nature but also enough images where the annotation is still uncertain. This is also visible in the majority votes. While each class contains originally 1000 images, the number of images classified into the categories according to the majority vote varies slightly between 981 and 1015. Additionally, the images can be easily assessed and evaluated against the annotations, in contrast to the dataset presented previously. By applying the proposed model we generate embeddings of the images in the appropriate label space, which contain a notion of uncertainty and reflect the original annotations, without the loss of information by taking the majority voting.

\begin{figure}[h]
\centering
\begin{minipage}{0.25\linewidth}
\flushright
\includegraphics[]{plots/cifar/404_ship.png}
\end{minipage}
\begin{minipage}{0.74\linewidth}
    \includegraphics[width=0.7\textwidth]{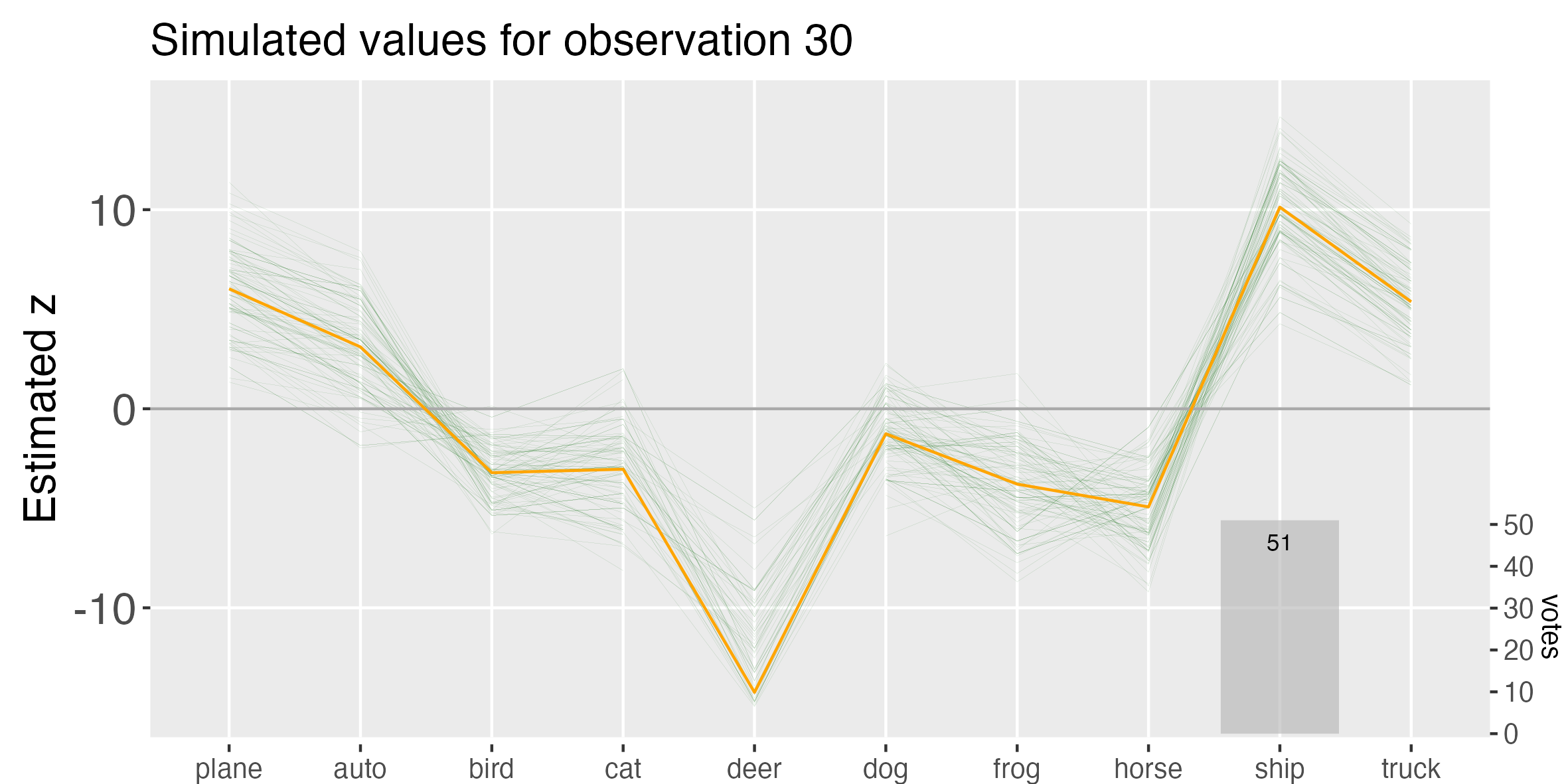} 
\end{minipage}

\begin{minipage}{0.25\linewidth}
\flushright
\includegraphics[]{plots/cifar/5471_frog.png}
\end{minipage}
\begin{minipage}{0.74\linewidth}
    \includegraphics[width=0.7\textwidth]{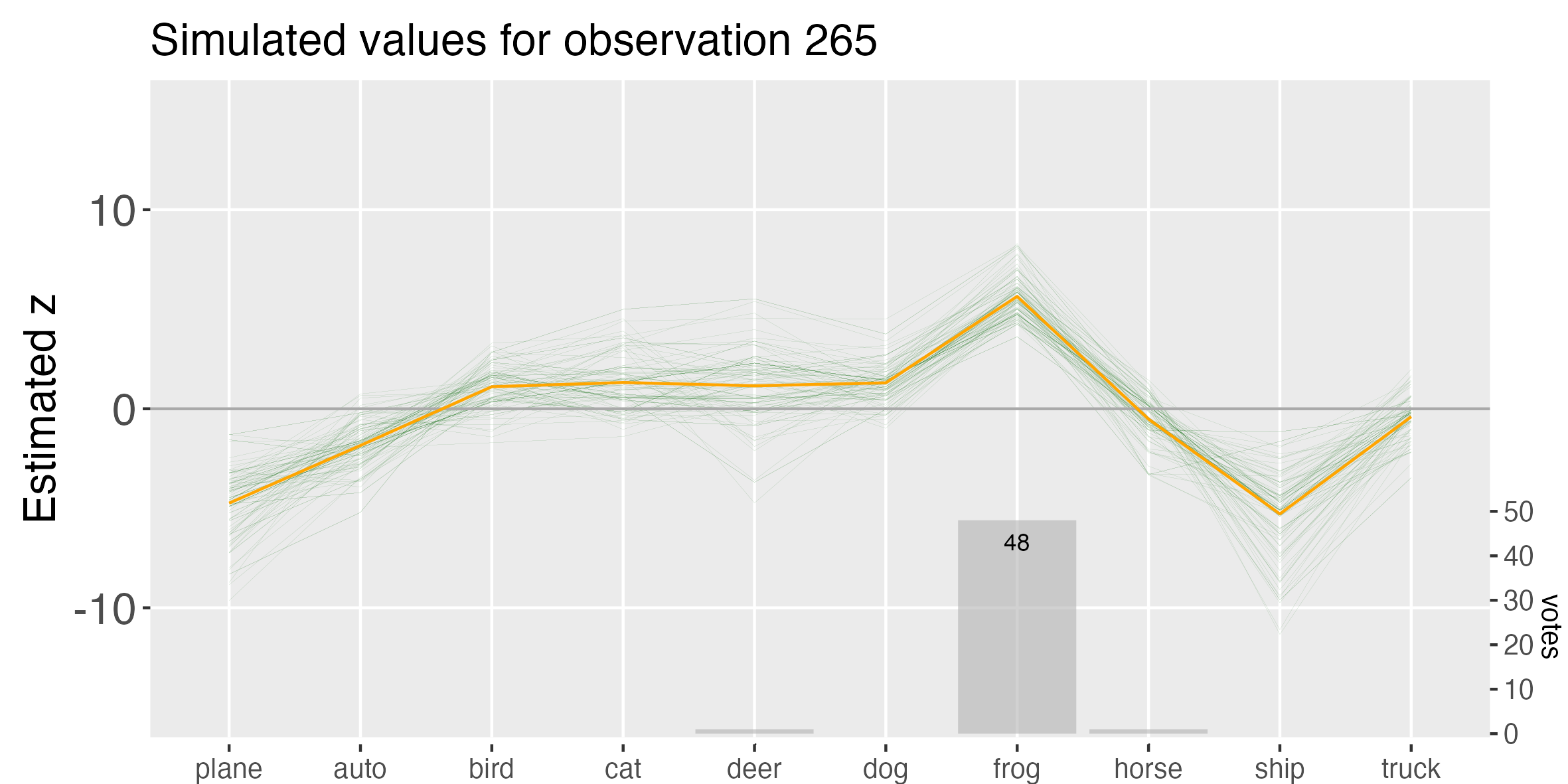}
\end{minipage}

\begin{minipage}{0.25\linewidth}
\flushright
\includegraphics[]{plots/cifar/3463_cat.png}
\end{minipage}
\begin{minipage}{0.74\linewidth}
    \includegraphics[width=0.7\textwidth]{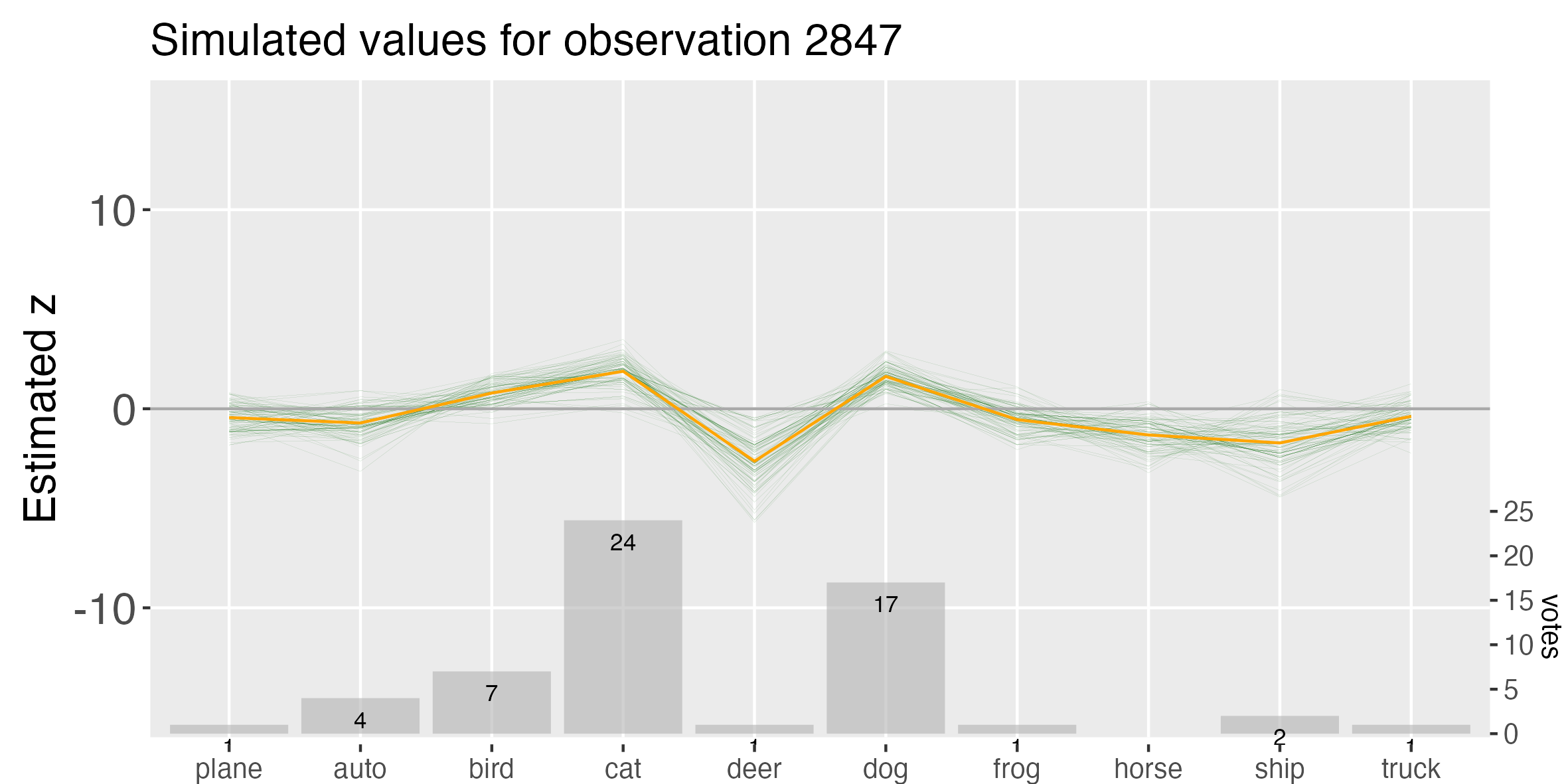}
\end{minipage}

\begin{minipage}{0.25\linewidth}
\flushright
\includegraphics[]{plots/cifar/6750_deer.png}
\end{minipage}
\begin{minipage}{0.74\linewidth}
    \includegraphics[width=0.7\textwidth]{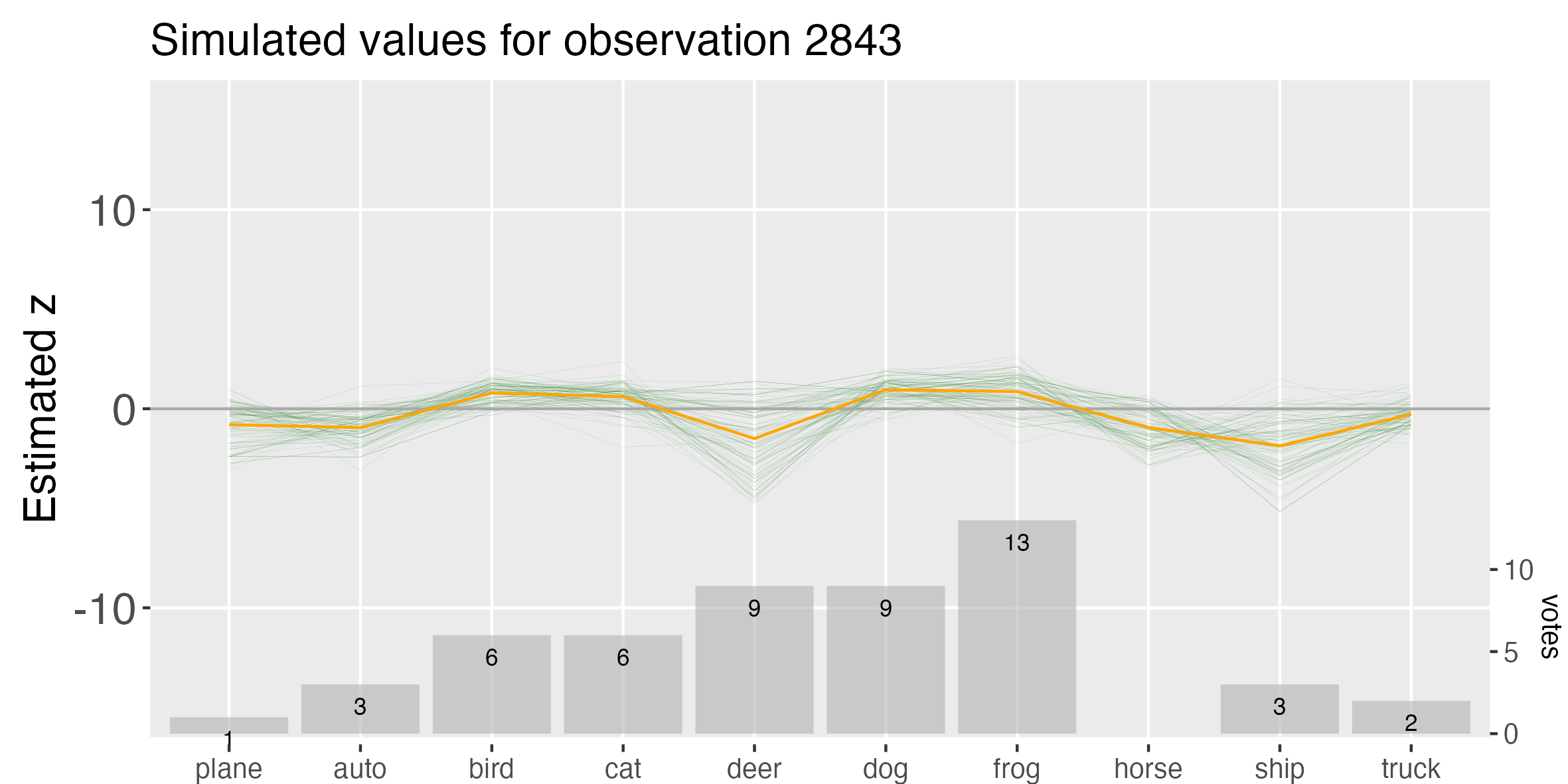} 
\end{minipage}
    \caption{The plots show the estimated embeddings (orange) for exemplary images of the dataset Cifar-10H, along with the votes (bars) and the MCMC samples (green).}
    \label{fig:cifar_example_embeddings}
\end{figure}

Returning to the \textbf{exemplary images} from Figure \ref{fig:intro_examples}, the estimated ground truth embeddings are shown in Figure \ref{fig:cifar_example_embeddings}. The upper plot in Figure \ref{fig:cifar_example_embeddings} shows the label embedding for an image of a ship, which is clearly visible in the picture and therefore also identifiable by the annotators. This is reflected in the respective label embedding with a high positive value for class \textsl{ship}.
For the second image, two annotators did not agree with the others and labeled the picture of a frog as \textsl{horse} or \textsl{deer}. Most of the labelers could correctly assign the label \textsl{frog}. The estimated label embedding reflects this by assigning the highest value to the class \textsl{frog} and small positive values to the other two classes. Nevertheless, the classification of the image is apparently easy, which is expressed by the embedding.
This does not hold for the images, which correspond to the two lower plots. The correct label for the third image was \textsl{cat}, correctly identified by the majority vote. Nevertheless, the image is quite ambiguous, which is reflected in the annotations and therefore also in the label embedding. Only using the majority vote would therefore lead to a correct label while losing a large amount of information about the inherent uncertainty. 
For the last image, the annotators did not agree at all and could not recover the label \textsl{deer}. In fact, almost every class received votes. In this case, the label embedding clearly reflects this confusion by assigning similar values close to zero to all classes, reflecting the classification uncertainty. 

\begin{figure}
    \centering
\begin{subfigure}{0.49\textwidth}
    \includegraphics[width=\textwidth]{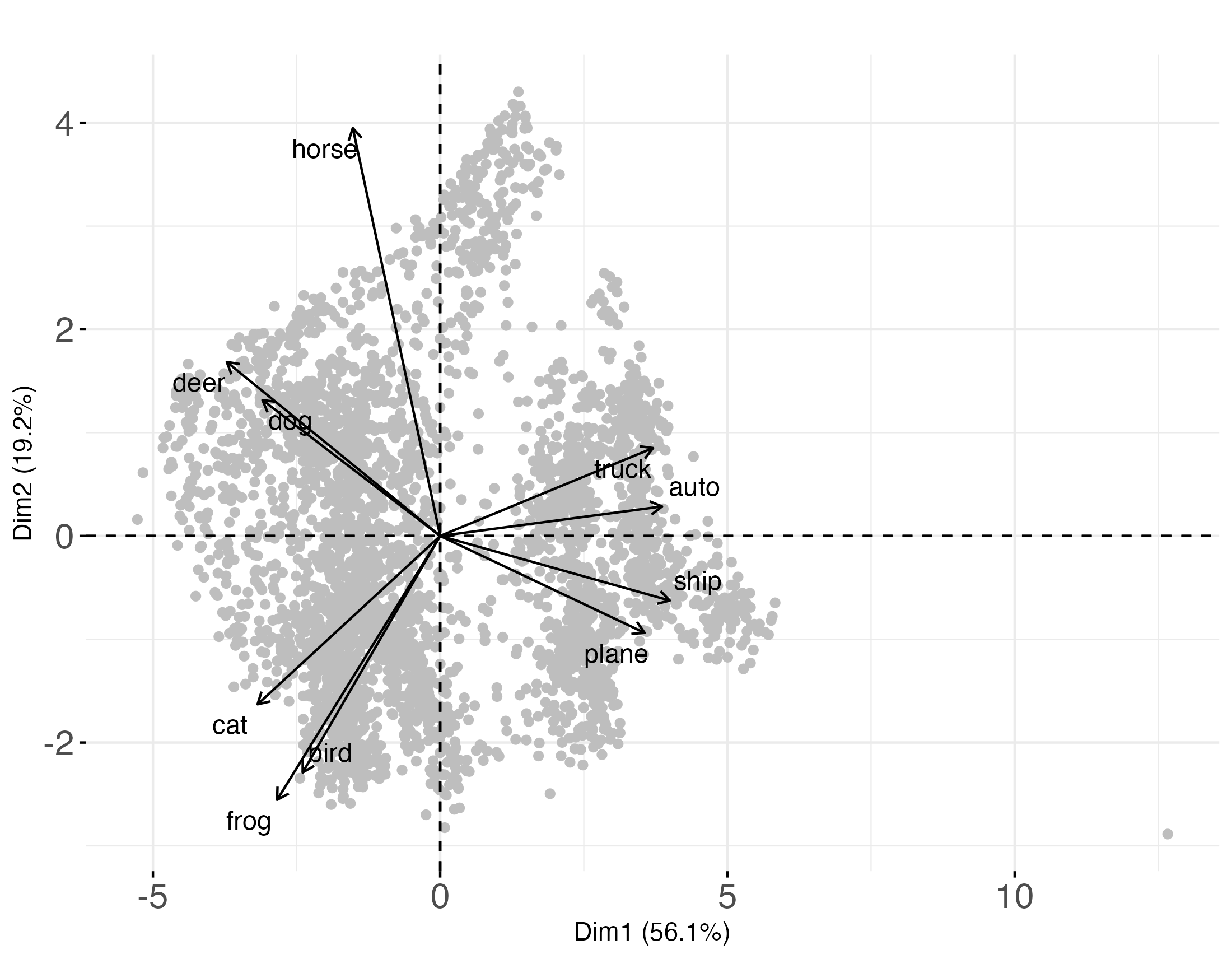}
    \caption{Biplot.}
    \label{fig:biplot_cifar}
\end{subfigure}
\begin{subfigure}{0.49\textwidth}
    \includegraphics[width=\textwidth]{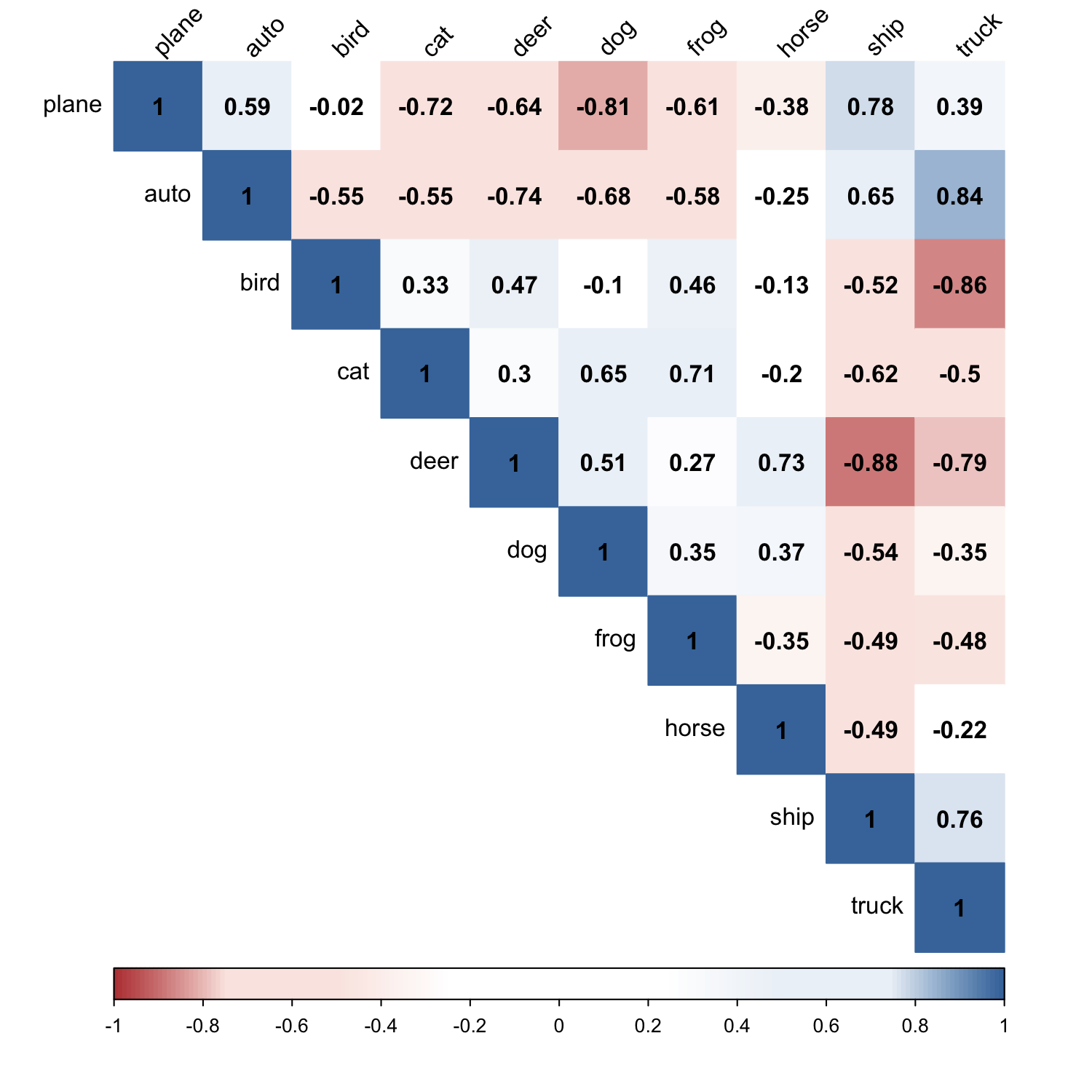}
    \caption{Correlation Matrix.}
    \label{fig:corr_cifar}
\end{subfigure}
\caption{The subfigures show additional results for the dataset Cifar-10H via the biplot and the correlation matrix of the estimated embeddings.}
\label{fig:cifar}
\end{figure}

Next, we repeat the analyses for the previous datasets, i.e. plotting the estimates of the embeddings onto a 2-dimensional \textbf{biplot} via PCA as well as calculating their \textbf{correlation matrix}.
The biplot of the projected embeddings is displayed in Figure \ref{fig:biplot_cifar} and shows a clear separation between classes referring to animals and classes referring to objects.
This is also expressed by the correlation matrix, shown in Figure \ref{fig:corr_cifar}. Again, this reflects possible similarities between images from correlated classes, which occur due to individual human perception despite the clear separation of the classes by definition. 

\section{Outlook}

Naturally, the question arises of how to use the information gained from embedding the ambiguous labels into a multidimensional space. 
The two main goals are to improve the supervised model assigned with the corresponding classification task and to possibly refine its uncertainty estimates. 
In many applications, training the classification model based on averaged labels or labels obtained via majority voting is still common practice. In the case of high annotator disagreement due to ambiguities, this can lead to major problems related to the associated uncertainties (\citealp{plank:2022}, \citealp{baan:2023} and \citealp{davani:2022}). \cite{koller:2024} propose to instead integrate the annotation uncertainty via the empirical distribution of the annotations. Their work shows that incorporating this uncertainty leads to better generalization and calibration of the classification model. 
However, the benefit of the empirical distribution of course strongly depends on the number of annotations and is limited to the observed disagreement for one single instance only. 
The idea of estimating label embeddings via a distributional approach presented in this work offers a possibility to overcome said limitations.
In particular, it is possible to train a classification model directly on the estimated embeddings $ \hat{\boldsymbol{z}} = (\hat{\boldsymbol{z}}_1,...,\hat{\boldsymbol{z}}_n)$, resulting from the estimation process as the mean of the MCMC samples in the last iteration. These embedded ground truth values retain information about all annotations for the respective observation and additionally incorporate knowledge about the annotations globally, across all instances. 
This leads to a more sound representation of the labels expressing uncertainties due to ambiguities of the instances themselves and also ambiguities due to the similarity of specific categories. Hence, this approach naturally handles images that cannot be directly classified into one class only. 
To integrate the embeddings into a deep learning framework, several strategies are available. While it is possible to directly learn the embedded ground truth vectors via a regression framework, reformulating the label embeddings into a Dirichlet function also allows us to stay within the world of classification. 
Either way, by incorporating the embeddings as labels in a machine learning framework, we expect the model to be better calibrated and yield more expressive predictive uncertainties. 
While this is beyond the scope of this paper, the results presented here serve as a valuable starting point for future work.

\section{Discussion}

For classification models, the dependence on labeled training data is a common practice, i.e.\ each instance is linked to an established ground truth label or ``gold'' label. Generating these ground truth labels requires substantial human effort and is prone to errors causing uncertainty. However, unreliable labels cannot always be attributed to human failure. In many applications, assigning a single label is unrealistic or even impossible due to the ambiguity of the instances themselves. A single ground truth label often cannot account for the complexity of e.g. images or sentences. This is often expressed through a high rate of disagreement in the annotations received from human labelers. 
Hence, the single-label approach results in a substantial loss of information and introduces additional uncertainty into the classification process. Therefore, moving beyond this limiting assumption is necessary in certain applications. This can be done by considering more flexible and adaptive strategies. \\
This paper focuses on classifying text or images addressing the specific case where we cannot assume that every observation can be uniquely classified into one class.  
Based on multiple annotations per observation, we propose to embed the images into a $K$-dimensional space instead of restricting them to a single label.


The proposed estimation procedure leads to interesting results, as reported in Section 3. We estimate label embeddings for three different datasets, emphasizing the generality of our approach and its usefulness in diverse settings. 
First, we apply the method to the dataset ChaosSNLI from the domain of language classification. The dataset contains especially ambiguous sentence pairs and a high number of annotations per instance. The assumption of a singular gold label is especially doubtful for the classification of language, due to its inherent ambiguity and subjectivity. Instead, multi-dimensional embeddings can serve as a more appropriate representation of the underlying truth. We show that the estimated vectors not only express the uncertainty associated with the instances due to the observed annotator disagreement but also the uncertainty due to a possibly insufficient number of annotations. 
Second, we move on to the domain of image classification and apply the proposed method to the earth observation dataset So2Sat LCZ42. Here, the satellite images themselves exhibit a high degree of ambiguity but also the categories are similar in terms of their composition, complicating the assignment of a singular label even more. 
Therefore, we inspect the correlation matrix based on the estimated embeddings. This matrix can be interpreted as a generalized confusion matrix, where a high correlation refers to a high confusion risk and vice versa. The correlation values directly reflect the semantic similarities of the LCZs. A small fraction of uncertainty remains as we only inspect a limited number of images. The chosen model framework allows estimating a distribution of the embeddings, namely a multivariate Gaussian, where the variance matrix expresses the uncertainty. 
Third, we move away from expert labels and inspect the performance of our model on a crowd-sourced dataset, namely the multiply annotated dataset Cifar-10H. The results show that even the classification of images into well-separated and naturally distinguishable categories could benefit from using label embeddings instead of hard-coded labels. 
The proposed model and the estimation framework are very flexible and hence, the presented work can be easily adapted to any classification problem with multiple annotations.

These insights can be valuable in multiple regards and pave the way for future research in various directions. While the presented results already deliver interesting insights into the annotation tasks, they of course rather serve as a preprocessing step for further work. 
The long-term goal is to use label embeddings within a complete machine-learning framework. In particular, we are interested in training classification models on multi-dimensional embeddings instead of single labels, i.e.\ incorporating information about label uncertainty directly into the model. 
This work can also serve as a basis for analyzing different design choices for label generation for image classification problems. The trade-off between the number of instances and the number of annotators is a well-known problem, related to experimental design. For problems with a high degree of ambiguity, determined by the proposed model, acquiring more annotations instead of more instances is beneficial. Vice versa, if classification is ``easy'', i.e.\ the embeddings reflect clear class affiliations, a smaller number of annotations might be sufficient and one should concentrate on generating more labeled instances instead. 
We believe that our modeling framework could be of great benefit for future steps towards better handling label uncertainty for machine learning models.

\section*{Acknowledgements}
The present contribution is supported by the Helmholtz Association under the joint research school “HIDSS-006 - Munich School for Data Science@Helmholtz, TUM\&LMU".

\bibliographystyle{plainnat}
\bibliography{bibliography} 

\newpage

\appendix

\section{Implementation Details}
\label{sec:appendix-algorithm}
The algorithm presented in Figure \ref{fig:mcmc_embedding} was initialized with 
\begin{align*}
    &\mu^{(1)} = (0,...,0) \in R^K \\
    &\Sigma^{(1)} = \begin{pmatrix}
10 & 0 & ... & 0 & 0 \\
0 & 10 & ... & 0 & 0 \\
\vdots & \vdots  & \ddots & \vdots  &\vdots \\
0 & 0 & ...  & 10 & 0 \\
0 & 0 & ... & 0 & 10
\end{pmatrix} \in R^{K \times K}.
\end{align*}
Specifically, we employed the Metropolis Hasting algorithm implemented in the function \texttt{MCMCmetrop1R()} from the R package \textbf{MCMCpack} (\citealp{mcmcpack:2011}) for drawing MCMC samples ($\texttt{mcmc}=1000$) from the logarithmic version of the posterior. Thereby, $\texttt{burnin}=50$ samples were discarded during the burn-in phase and $\texttt{thin}=20$ was used as thinning interval.
The starting value $\texttt{theta.init}$ was set to $\hat{z}_i^{(m-1)}$, i.e. the previously estimated embedding vector, or to $(0,...,0)$ in the first iteration.
Additional details and the code can be found on github via \texttt{https://github.com/katharinahech/labelembeddings}

\section{ChaosSNLI: Majority Vote vs. Ground Truth}
\label{sec:appendix:m-vote}

\begin{figure}
    \centering
    \includegraphics[width=0.49\textwidth]{plots/chaosnli/biplot_majority_vote.png}
    \includegraphics[width=0.49\textwidth]{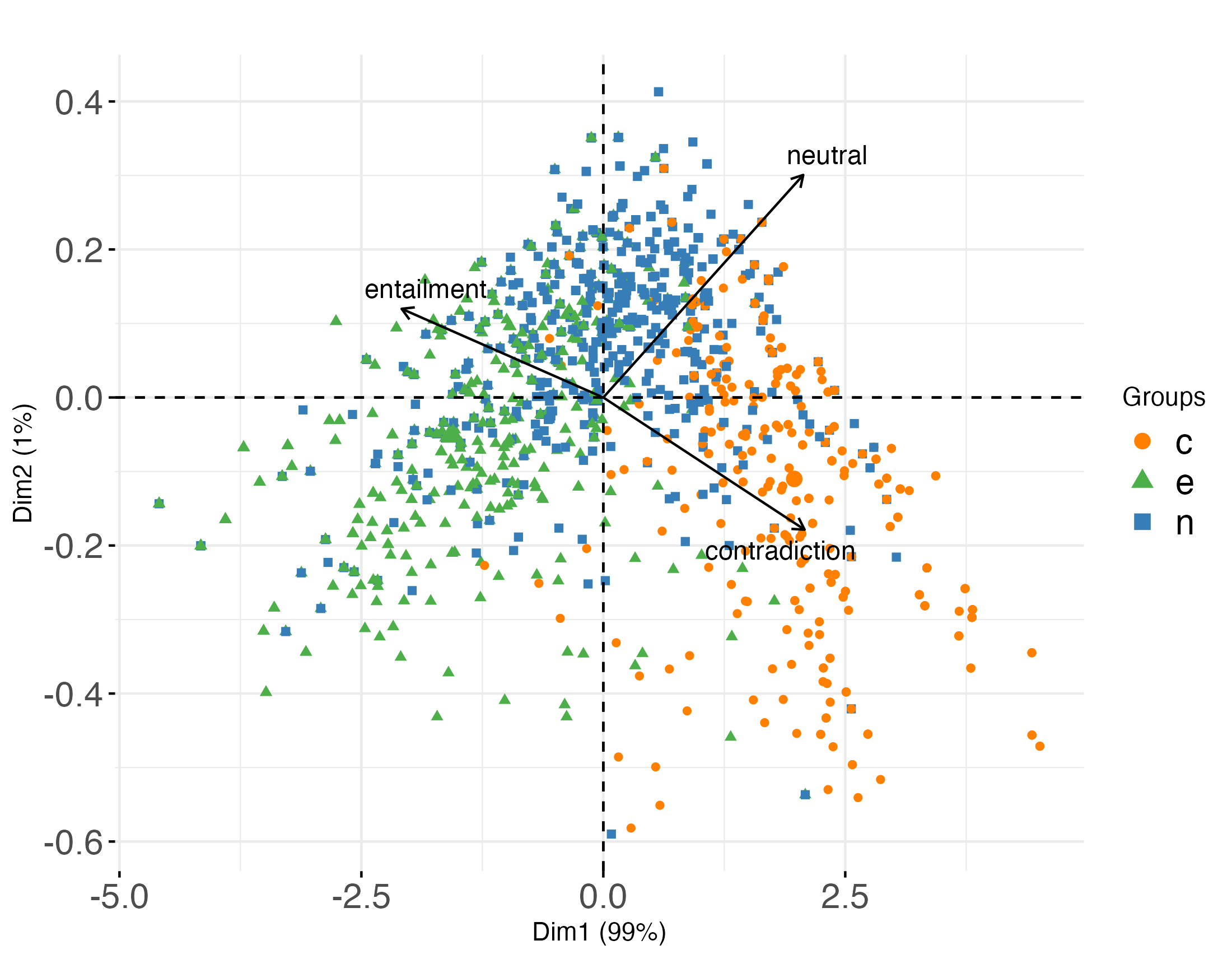}
    \caption{The biplots show the projected embeddings, colored by majority vote (left) and by original subjective ground truth label (right).}
    \label{fig:appendix-gt}
\end{figure}

The dataset ChaosSNLI is especially interesting in the context of analyzing label variation and ambiguities. It does not only contain a huge number of annotations but also provides a subjective ground truth label for each pair of sentences due to its generation process, see \cite{nie:2020} for details. However, due to the general ambiguity of language, the subjective "true" label does not constitute a reliable ground truth in general. Nevertheless, it is interesting to examine the difference between the original label and the majority-voted label, as shown in Figure \ref{fig:appendix-gt}. While the left plot shows the biplot of the projected embeddings colored by the majority vote, the points in the right plot are colored according to the original "gold" label. For the latter, we see less separation of the three classes in the two-dimensional space. In particular, instances with the original label \textsl{neutral} are often classified into \textsl{contradiction} or \textsl{entailment} according to the majority vote. This observation suggests that the class "neutral" is harder to detect as the other classes are inherently more informative, see also \cite{gruber:2024}.

\section{ChaosSNLI: Impact of the Number of Annotations}
\label{sec:appendix-annotations}

\begin{figure}
    \centering
    \includegraphics[width=0.5\textwidth]{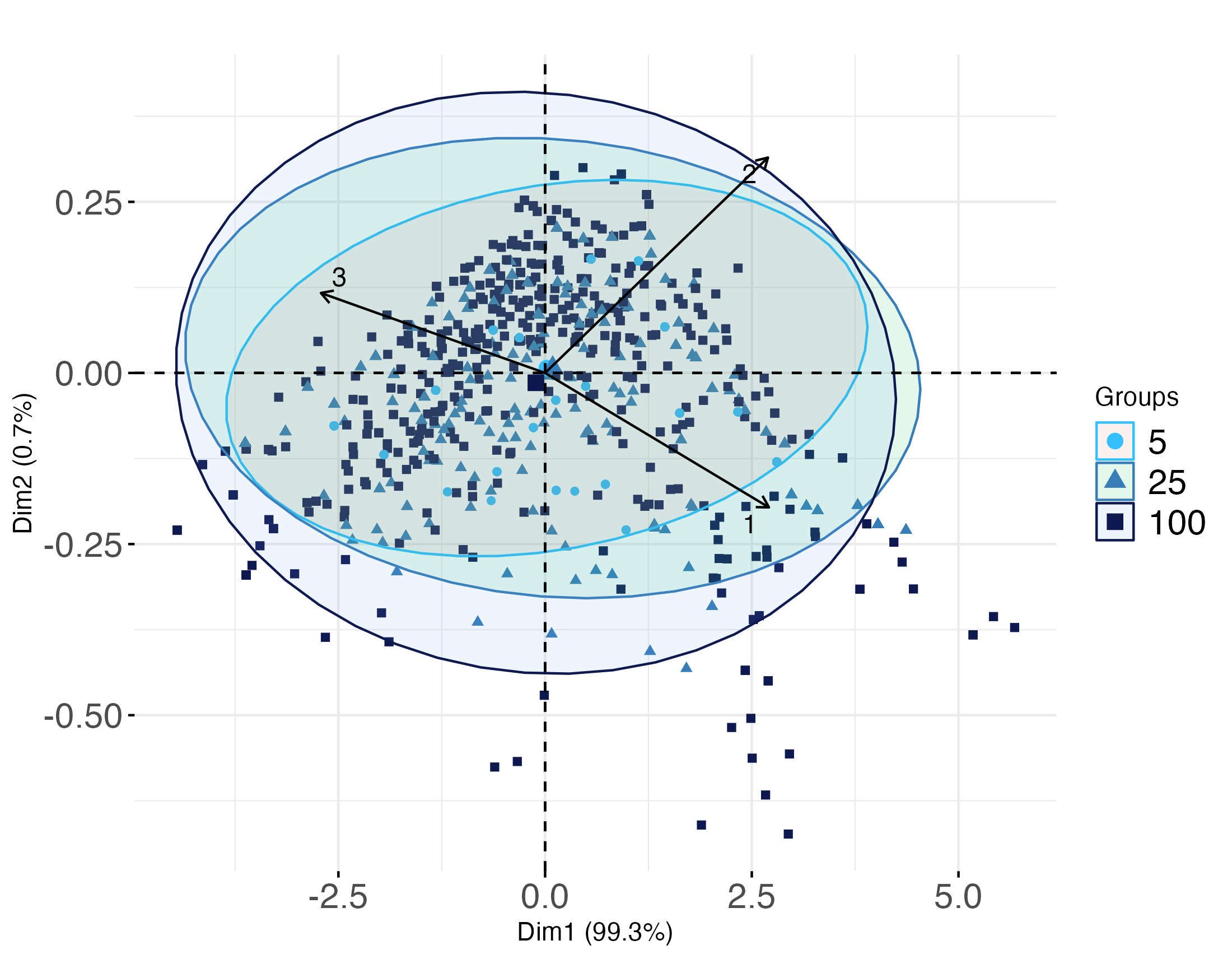}
    \caption{The biplot shows the projected estimated embeddings for a version of the dataset ChaosNLI, where for random instances 5, 25 or 100 annotations were sampled. The points are colored according to the number of annotations and concentration ellipses are shown for easier visual exploration.}
    \label{fig:biplot_chaosnli_annotations}
\end{figure}

The dimensions of the dataset also allow us to investigate the dependence of the results on the number of annotations, $J$. Following the sampling strategy proposed by \cite{gruber:2024}, we randomly sample annotations for a fraction of the original dataset and thereby create a new version, consisting of observations with a varying number of annotations. Specifically, $J_{100} = 100$ annotations remain for $N_{100}=514$ observations, while for $N_5=500$ resp. $N_{25}=500$ instances $J_{5}=5$ resp. $J_{25}=25$ annotations were randomly sampled.
Figure \ref{fig:biplot_chaosnli_annotations} shows the biplot created based on the estimated ground truth embeddings for the sampled dataset.
The plot inherently expresses the variance of the instances, comparable to Figure \ref{fig:meanbetabin}.
Points located close to the center of the graphic exhibit a high degree of variance. In contrast, instances that lie far from the origin have a smaller variance, as shown for $K=2$ in Figure \ref{fig:meanbetabin}. Interestingly, Figure \ref{fig:biplot_chaosnli_annotations} clearly reveals that observations with fewer annotations lie closer to the center of the plot compared to similar instances with a higher number of annotations, indicating higher variance and higher uncertainty. This is of course a highly desirable property of the embedded ground truth vectors. The estimated values express both, the uncertainty due to the disagreement of the labelers as well as the uncertainty due to an insufficient number of annotations. Figure \ref{fig:biplot_chaosnli_annotations} additionally hints at the fact that a certain number of annotations is necessary to fully capture the associated ambiguity (\citealp{gruber:2024}). The concentration ellipses for $J_{100}$ and $J_{25}$ are almost similar, whereas the ellipse for $J_5$ is notably smaller, indicating a higher variance of instances with only five annotations. 

\section{So2Sat LCZ42: Standard Deviation of the Correlation Matrix}
\label{sec:appendix-std}
\begin{figure}[h]
    \centering
    \includegraphics[width=0.7\textwidth]{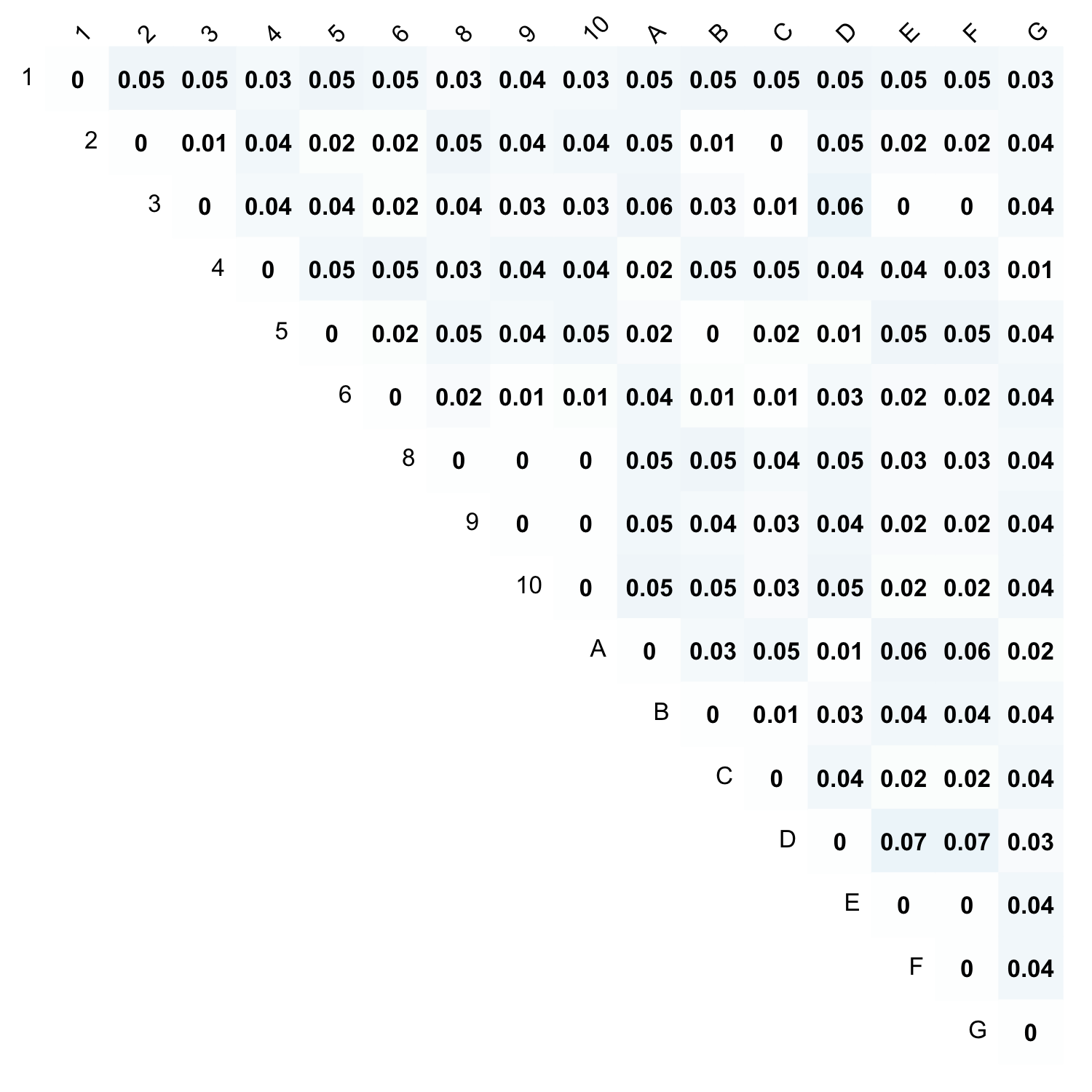}
    \caption{Standard deviation of the entries of the correlation matrix.}
    \label{fig:corr_so2sat_std}
\end{figure}

As mentioned in Section 3, the MCMC samples enable us to additionally inspect the variance of the correlations of the estimated embeddings. Figure \ref{fig:corr_so2sat_std} explicitly shows the respective variances of entries of the correlation matrix given in Figure \ref{fig:corr_so2sat}. 

\end{document}